\documentclass{article}
\usepackage{hyperref}
\usepackage{cite}
\usepackage{xcolor}
\usepackage{amsmath,amssymb,amsfonts}
\usepackage{algorithmic}
\usepackage{graphicx}
\usepackage{textcomp}

\hypersetup{
	colorlinks,
	citecolor=blue,
	linkcolor=red,
	urlcolor=blue,
  citebordercolor=blue,
filebordercolor=red,
linkbordercolor=blue
}

\usepackage{framed,multirow}
\usepackage{makecell}
\usepackage{latexsym}
\usepackage{amsmath}
\usepackage{subfigure} 
\usepackage{bbm}

\def\BibTeX{{\rm B\kern-.05em{\sc i\kern-.025em b}\kern-.08em
    T\kern-.1667em\lower.7ex\hbox{E}\kern-.125emX}}
\begin{document}

\title{Disentangled representations: towards interpretation of sex determination from hip bone}
\author{Kaifeng Zou, Sylvain  Faisan, Fabrice Heitz\thanks{K. Zou,  S. Faisan and F. Heitz are with ICube laboratory of university of Strasbourg, Strasbourg, France (email: kaifeng.zou@unistra.fr, faisan@unistra.fr, fabrice.heitz@unistra.fr)}
, Marie Epain,\\ Pierre Croisille\thanks{P. Croisille is with University Hospital of Saint-Etienne, France}, Laurent Fanton\thanks{M. Epain and L. Fanton are with Hospices Civils de Lyon; France (email : firstname.lastname@chu-lyon.fr).} , Sébastien Valette \thanks{S. Valette, P. Croisille and L. Fanton are with CREATIS, INSA-Lyon, Lyon, France (email: firstname.lastname@creatis.insa-lyon.fr).} \thanks{This work was funded by the TOPACS ANR-19-CE45-0015 project of the French National Research Agency (ANR). \copyright 2021 IEEE. Personal use of this material is permitted. Permission from IEEE must be obtained for all other uses, in any current or future media, including reprinting/republishing this material for advertising or promotional purposes, creating new collective works, for resale or redistribution to servers or lists, or reuse of any copyrighted component of this work in other works.}}

\maketitle

\begin{abstract}
By highlighting the regions of the input image that contribute the most to the decision, saliency maps have become a popular method to make neural networks interpretable. In medical imaging, they are particularly well-suited to explain neural networks in the context of abnormality localization. However, from our experiments, they are less suited to classification problems where 
the features that allow to distinguish between the different classes are spatially correlated, scattered and definitely non-trivial.
In this paper we thus propose a new paradigm for better interpretability.
To this end we provide
the user with relevant and easily interpretable  information so that he can form his own opinion. We
use Disentangled Variational Auto-Encoders which latent representation is divided
into two components: the non-interpretable part and the disentangled
part. The latter accounts for the categorical variables explicitly representing the different
classes of interest. In addition to providing the class of a given input sample, such a model
offers the possibility  to transform the sample from a given class to a sample of another
class, by modifying the value of the categorical variables in the latent representation. This
paves the way to easier interpretation of class differences. We illustrate the relevance of
this approach in the context of automatic sex determination from hip bones in forensic
medicine. The features encoded by the model, that distinguish the different classes were
found to be consistent with expert knowledge.
\end{abstract}

\section{Introduction}
\label{sec:introduction}
Neural networks-based classification methods
are often criticized for their lack of interpretability and explainability. Interpretability and explainability aim to understand
what the neural network has learned or how prediction is performed. 
Saliency maps\cite{simonyan2014deep} are widely used in medical imaging to provide interpretability and explainability, especially in the context of abnormality localization.
However, in our experiments, the information extracted with saliency maps  was difficult to interpret. 
Our hypothesis is that saliency maps 
of neural networks 
are not easily interpretable on medical imaging classification problems where  
the underlying features used by the neural network to support the decision are spatially correlated, scattered and non-trivial.

To overcome this limitation, we consider here a different paradigm, based on disentangled generative representations, 
to help the interpretation of deep neural networks decision.

Probabilistic generative models, such as Variational Auto-Encoders (VAE) \cite{Kingma2014}, define a joint probability distribution over the data and over latent random variables. Very few assumptions are generally made on the latent variables of deep generative models, leading to entangled representations. For medical applications, it is often needed to identify standard sources of variability such as acquisition parameters, age, sex or pathology. There is thus a key challenge to learn disentangled representations where variables of interest would be independently and explicitly encoded \cite{bengio}. 
There are three main paradigms to learn disentangled representations: unsupervised \cite{unsupervised,mic2}, supervised or semi-supervised \cite{Siddhart,KingmaD,mic}, and weakly-supervised \cite{ruiz2019learning}.
In the supervised or semi-supervised case, the factors of interest are explicitly labelled in all or in a part of the training set. In the weakly-supervised case, only implicit information about factors of interest is provided during learning. 
In the following, we are interested in the supervised case.
Disentangled representations allow to reveal the effects of the factors of interest, through the generation of new data, by changing value-level factors \cite{YanYSL16}. As an example, \cite{mic2} samples the latent space so as to provide insights from brain structure representations. 
Another model proposed in \cite{mic} can simulate brain images at different ages, providing an alternative way for interpreting the aging pattern. 

We show in this paper that disentanglement also brings a better understanding of classification results, highlighting the differences between the possible classes. We illustrate this ability in the context of sex determination from the hip bone. 
In forensic medicine and anthropology, sex determination is generally carried out by manually assessing hip bone features \cite{forensincsAnthropology}.
Automatic classification algorithms are mainly guided by the knowledge of anthropologists, taking into account distances or angles measured on a few anatomical landmarks \cite{CADOES2019,MURA2005,BRUZ2017,Nikita2020}.
Such approaches have the advantage of providing easily interpretable results. But they sometimes fail and are specifically tailored for hip bones, hence not well suited to the classification of other bones or bone fragments, which may be necessary in forensic science.
In  contrast, our approach is completely data-driven and free of expert knowledge, while providing interpretability and explanability to the practitioner. It is also suited to the classification from other bones or bone fragments, while dealing with missing data.

We introduce a disentangled Variational Auto-Encoder (DVAE) for hip bone meshes representation, that disentangles the sex label of interest from the other latent random variables. In addition to providing the class of a given sample to analyze, a DVAE can also provide a reconstruction for each class, which brings more information to the user.
As an example, if the input mesh is a male one, its reconstruction as a man should be similar to the original mesh. Conversely, the reconstruction as a woman should exhibit interpretable differences in sex-specific regions. Moreover, by comparing the two reconstructions with the original mesh for several subjects, the user can get an insight into the morphological differences between male and female hip bones. We also show in this paper that feeding a binary classifier with these two reconstructions increases accuracy even further.

Even if there is not a clear consensus on the definition of interpretability and explainability, most methods aim to understand what the neural network has learned or how prediction is done. 
In the literature, there are mainly two approaches to provide this interpretability and explainability. The first paradigm, known as activation maximization or feature visualization
via optimization, consists in  producing intuitive visualizations that reveal the meaning of hidden layers. This is mainly achieved by finding a representative input that can maximize the activation of a layer \cite{vis,Nguyen2019}. 
The second paradigm, known as attribution methods, looks for the network inputs with the highest impact on the network response. In the case of image models, this leads to the estimation of saliency maps (SM), which highlight the regions of the input image that contribute the most to the decision.
Many attribution techniques are based on backpropagation. A SM is for instance computed in \cite{simonyan2014deep} by computing the derivative of the output with respect to the image. Several methods such as SmoothGrad \cite{smoothgrad} have been proposed to reduce the noise that is present in the gradient. Methods such as  CAM\cite{CAM2} and Grad-CAM\cite{Selvaraju_2019} combine gradients, network weights and/or activations at a specific layer. Other attribution techniques analyze how a perturbation in the input affects the output \cite{9010039}. Finally, attribution techniques can also be achieved via local model approximation \cite{LIME}. 
In the context of medical imaging, SM are now a popular approach that provide interpretability, especially in the context of abnormality localization.
Different sanity checks \cite{arun2020assessing}, such as intra-architecture repeatability,  inter-architecture reproductibility, sensitivity to weight randomization \cite{sanityCheck} and localization accuracy can be used to assess the relevance of SM.
These criteria helped to justify the use of SM in some studies such as in \cite{arun2020assessing}, but have also led to questions about the relevance of SM\cite{eitel2019testing,three}.
This indicates that SM are not suited for all situations. As already stated, in our application, 
information extracted with SM was difficult to interpret (examples of SM are presented on Fig. \ref{fig:saliency}).

Note also that an intrisic limitation of SM is that they do not provide any semantic meaning on the highlighted regions. In our application, the SM can at best detect sex-specific regions, i.e., regions that allow to distinguish between male and female hip bones. In contrast, by reconstructing a hip bone either as a male or a female one, the proposed method reveals {\em all} the differences between men and women hip bones: the proposed method not only provides a sex-specific region detection but also offers the user the opportunity to observe the difference in shape of regions.
Such an approach  leads to a better understanding of the class differences.

Note finally that the proposed approach is only suitable if the label to estimate is a  variable corresponding to a source of variability (age, sex, outcomes of genomic-biological-cognitive tests, diagnosis, multicenter variability, ...), which are common situations in medical imaging.  As an example,  it makes sense in the proposed application to reconstruct a male hip bone as a female one (or a diseased organ into a healthy one, ...) because the latent space can be divided into two independents parts: the non-interpretable part represents the intrinsic (independent of sex) properties  of the hip bone and the disentangled part represents the sex label.

The remainder of this paper is organized as follows: in Sec. \ref{sec:preprocessing}, 
we briefly explain how hip bone meshes are obtained from CT scans. 
Sec. \ref{sec:method} presents the DVAE. Sec. \ref{sec:res} presents  the experiments  as well as the results and Sec. \ref{sec:dis} proposes a discussion.
If the two reconstructions provided by the DVAE enable the 
user to form his own opinion, Sec. \ref{cls} shows that the two reconstructions may also be useful to improve the accuracy of an independent classifier. This section also addresses the case of missing data. In Sec. \ref{sec:saliency}, we provide saliency maps for the proposed networks for comparison.
Finally, Sec. \ref{sec:con} concludes the paper.

\section{From CT scans to meshes}
\label{sec:preprocessing}
In this section we assume that we have one 3D CT scan $I_k$ for each individual $k$.
Computing a mesh of the hip bone from a CT image is carried out in six steps: \emph{(i)} The scans are registered to a common space using the groupwise registration algorithm FROG \cite{AGIE2020}. This results in a transformation field $t_k$ that transforms each scan $I_k$ to the  reference common space. 
\emph{(ii)} Each scan $I_k$ is transformed in the common space,  
and a template average image denoted $T$ is computed. \emph{(iii)} The coxal bone is segmented and meshed in $T$, thus providing a mesh $M$ of the coxal bone. The mesh is composed of about 5000 vertices (we denote by $P$ the 3-$D$ points associated to the mesh $M$). 
\emph{(iv)} The points $P$ are back-transformed in the original space of each scan $I_k$ using the inverse transform $t_k^{-1}$, providing
for each scan $I_k$ a matrix $X_k$ of size $N_p \times 3$ ($N_p$ is the number of points). Each column of $X_k$ is the 3-D coordinate of one point. Note that points are ordered since the $i$-th column of each matrix is associated to the same ``anatomical" point.
\emph{(v)} A shape description invariant to position, size and orientation denoted $P_k$ is obtained using a Procrustes alignment of $X_k$ onto  $P$ (for each $X_k$, we estimate a similarity transformation, namely the combination of a rigid transformation with isotropic scaling transform). 
A shape description invariant to position and orientation is required since all subjects do not have the same position during acquisition whereas a description invariant to size is more debatable.
Finally, \emph{(vi)} since the point sets $P_k$ and $P$ are ordered, the mesh $M_k$ is straightforwardly derived from $M$ and $P_k$.

\begin{figure*}[t!]
\centering
\includegraphics[width=0.95\textwidth]{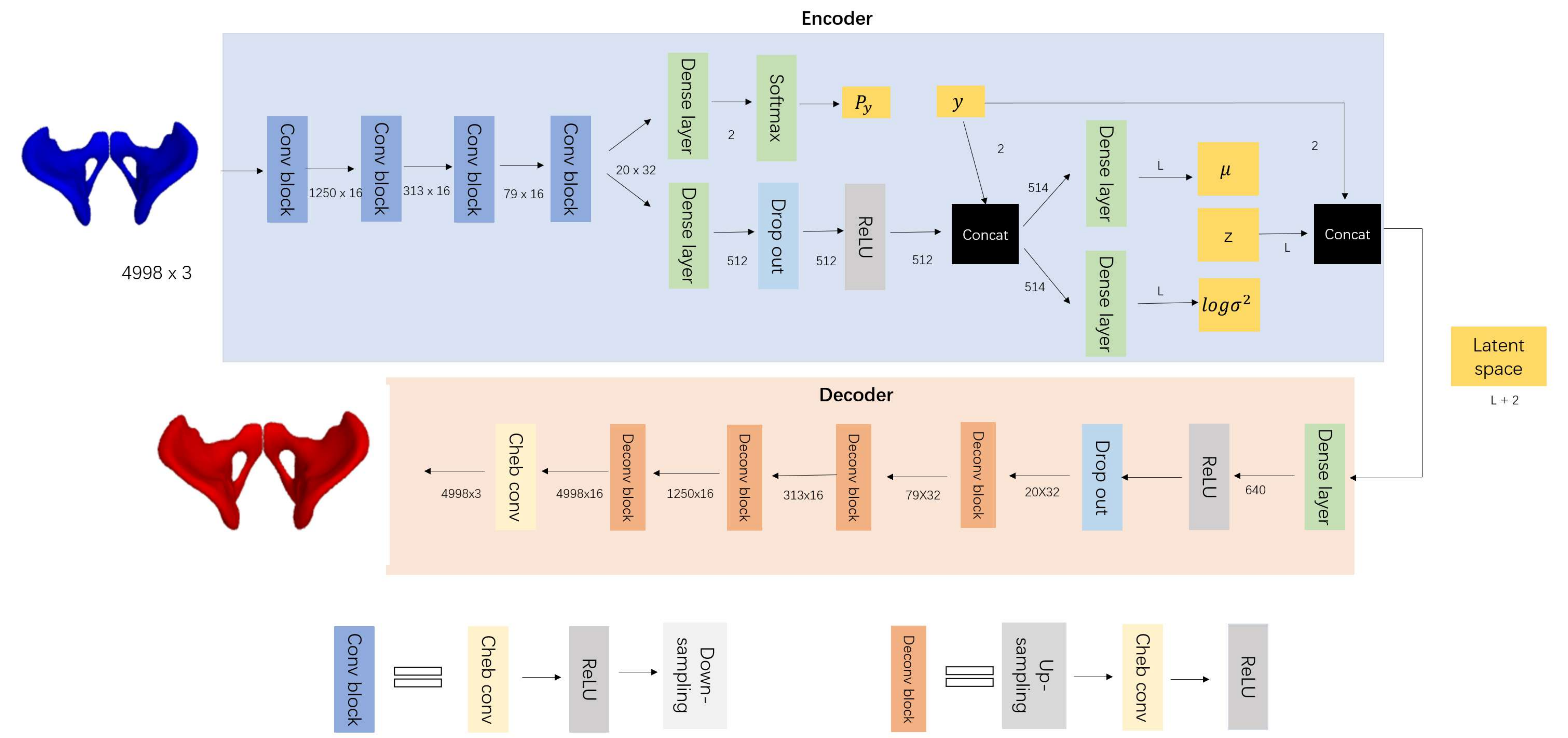}
\caption{DVAE for sex determination. Since hot encoding is used to model $y$, $y$ is of dimension 2. For learning, $y$ is known and the reparameterization trick
(Eq. \ref{eq:ss3}) is used to set $z$.
For testing, $y$ is is set to the most likely 
label thanks to $P_y$ and $z$ is set to $\mu$.} \label{fig1}
\end{figure*}

\section{Disentangled Variational Auto-Encoders for classification and reconstruction}
\label{sec:method}

\subsection{Conditional dependency structure}
\label{sec:cds}
The proposed model is part of the family of partially-specified models because an explicit latent variable is defined (the sex of the subject) whereas the semantic of the other latent variables is undefined.
Several conditional dependency structures can be defined. As an example, \cite{mic} explicitly  conditions the latent variables $z$ on age $c$, such  that the conditional  distribution $p(z|c)$  captures  an age-specific  prior on  latent  representations. We propose here to use a conditional dependency structure presented in \cite{Siddhart,KingmaD}, which is suited to our problem. 

We denote by $x$ a sample (a mesh), by $y$ its class (male or female), and by $z \in \mathbb{R}^L$ its latent representation.  We use the following factorization for the generative process: 
\begin{equation}
p_{\theta}(x,y,z) = p_{\theta}(x|y,z) p(y) p(z), 
\label{eq:genProcess}
\end{equation}
where a weak prior is defined over $z$ and $y$ : $p(z) = \mathcal{N}(z|0,I)$ and $p(y) = \frac{1}{2}$. $p_{\theta}(x|y,z)$ is modelled as a Gaussian distribution whose mean is given by a neural network $f$ of parameter $\theta$ that takes as input $y$ and $z$. We have:
\begin{equation}
\begin{array}{lll}
p_{\theta}(x | y,z ; \theta) &=& 
\mathcal{N}\left(x \mid f(y,z ; \theta),  v I  \right), \\
&=& 
\mathcal{N}\left(x \mid \widehat{x},  v I  \right),
\end{array}
\label{eq:P1}
\end{equation}
where $v > 0$ is a hyperparameter and  $\widehat{x}$ is the reconstruction related to $y$ and $z$.

As usual in variational inference, the posterior $p_{\theta}(y,z|x)$ is approximated by  $q_{\phi}(y,z|x)$. In order to disentangle the label $y$ from the other latent variables $z$, we use the following factorization:
\begin{equation}
q_{\phi}(y,z|x)=  q_{\phi}(y|x) q_{\phi}(z|x,y).
\label{eq:app}
\end{equation}

The distribution $q_{\phi}(z|x,y)$ shows that the estimation of $z$ requires the data $x$, but also the label $y$. To understand why this is important, let us consider a toy example where $z$ is supposed to represent the size of the subject. If the sex label $y$ is well disentangled from the latent space $z$, $z$ should be an intrinsic measure that informs about the size of the subject. This means that its estimation somehow requires to regress out the influence of the label $y$: indeed, a woman who is 160 centimeters tall can be considered  as average height while a man of the same height can be considered as short, so that the value of $z$ associated to this woman has to be larger than the one related to this man (even if they have both the same height). Consequently, in order to obtain a disentangled representation, it seems important that $z$ depends both on $x$ and $y$. 

The distribution $q_{\phi}(z|x,y)$ in Eq. \ref{eq:app} is defined as a Gaussian distribution whose mean (resp. covariance matrix) are given  by a neural network $q_1$ (resp. $q_2$) of parameter $\phi_1$ (resp. $\phi_2$) that both take as input $x$ and $y$: 
\begin{equation}
q_{\phi}(z|x,y) = \mathcal{N}(z;\mu,I \times \sigma^2), 
\end{equation}
where $\mu$ and  $\log \sigma^2$ are 
vectors of size $L$ (please see Eq. \ref{eq:ss2} for details).
Finally, the distribution $q_{\phi}(y|x)$ that also appears in Eq. \ref{eq:app} is simply defined as: 
\begin{equation}
q_{\phi}(y|x) = Discrete(y|q_0(x;\phi_0)),
\label{eq:ch}
\end{equation}
where $q_0$ is a neural network of parameter $\phi_0$ that takes $x$ as input.
The output of this network is a positive vector $P_y$ of size 2 summing to 1: the probability $q_{\phi}(y=i|x)$ is the $i$-th element of $q_0(x;\phi_0)$ ($i$ = 1 or 2).

Eq. \ref{eq:s1} to \ref{eq:ss4} summarize the proposed model.
\begin{equation}
P_y = q_0(x;\phi_0),
\label{eq:s1}
\end{equation}
\begin{equation}
\mu = q_1(x,y;\phi_1), \log \sigma ^2 = q_2(x,y;\phi_2),
\label{eq:ss2}
\end{equation}
\begin{equation}
z = \mu +  \sigma \odot \epsilon \mbox{, where } \epsilon \sim \mathcal{N}(0,I),  
\label{eq:ss3}
\end{equation}
\begin{equation}
 \widehat{x} = f(z,y; \theta),
\label{eq:ss4}
\end{equation}
The neural networks
$q_0$ (Eq. \ref{eq:s1}), $q1$ and $q2$ (Eq. \ref{eq:ss2}) represent the encoder and 
$f$ is the decoder (Eq. \ref{eq:ss4}). 
If $y$ is known, the neural network $q_0$ is not required. 
Otherwise, it acts like a classifier: the distribution $q_{\phi}(y|x)$ (Eq. \ref{eq:ch}) is  computed using Eq. \ref{eq:s1} and $y$ is  set to the most likely label. Then, $y$ and $x$ are used to compute the latent representation $z$: firstly, $\mu$ and $\log \sigma ^2$ are computed using eq. \ref{eq:ss2}. Then, 
the latent representation $z$ is set to $\mu$ for testing new data whereas Eq. \ref{eq:ss3} represents the reparameterization trick that is used for learning (please see next section). Finally, the  reconstruction $\widehat{x}$ can be obtained from $y$ and $z$ using the decoder $f$ (Eq. \ref{eq:ss4}).

The proposed architecture is depicted in Fig. \ref{fig1}.
Networks $q_0$, $q_1$, $q_2$ and $f$ (Fig. \ref{fig1}) are defined using a combination of the convolutions, max-pooling (downsampling) and upsampling operators presented in  \cite{COMA:ECCV18}. Note that mesh convolution is performed in the spectral domain with a kernel parametrized as a Chebyshev polynomial of order K ($K$ is set to 6).

\subsection{Parameter optimization}
\label{sec:opt}
As usual for learning a VAE, the  parameters of the DVAE  are set so as to maximize the Evidence Lower BOund (ELBO). 
We can show that the term $q_{\phi}(y|x)$  does not contribute to the loss function 
because all labels $y$ are known during training.
Thus, maximizing the ELBO does not allow the estimation of $\phi_0$ (Eq. \ref{eq:s1}).
Consequently, following \cite{Siddhart,KingmaD}, we add  a classification loss $\alpha \log q_{\phi}(y|x)$ to the ELBO term.
The criterion writes:
\begin{equation}
E_{z \sim q_{\phi}(z|x,y)} \left[\log \frac{p_{\theta}\left(x, y,z\right)}{q_{\phi}(z \mid x,y)} \right]+ \alpha \log q_{\phi}(y|x).
\label{crt}
\end{equation}

Based on the conditional dependency structure of the model, Eq. \ref{crt} can be simplified as:
\begin{equation}
\begin{array}{lll}
  E_{z \sim q_{\phi}(z|x,y)} \left[log(p(z)) - log(q_{\phi}(z|x,y))  \right] & ~&+ \\
  E_{z \sim q_{\phi}(z|x,y)} \left[\log(p_{\theta}(x|y,z))  \right] &~ & + \\
  \log(p(y)) + \alpha \log q_{\phi}(y|x) &~ & 
  \end{array}
  \label{eq:crit}
\end{equation}

The first term may be expressed as a Kullback–Leibler divergence $(-KL((q_{\phi}(z|x,y)||p(z)))$ which can be computed analytically since the encoder model and prior are Gaussian. The second term is approximated by a Monte Carlo estimate:  we use the SGVB estimator and the reparameterization  trick \cite{Kingma2014} (Eq. \ref{eq:ss3}). The third term corresponds to the prior of the label $y$, that has been  set to 1/2. Finally, the last term is computed by the neural network $q_0$.

The loss function contains two hyperparameters: $\alpha$ that weights the contribution of the classification loss $\log q_{\phi}(y|x)$, and the variance $v$ (Eq. \ref{eq:P1}), which is used to compute the second term of Eq. \ref{eq:crit}. As in the VAE case, the variance $v$ weights the contribution of the mean squared error reconstruction (related to the second term of Eq. \ref{eq:crit}). A $\beta$-VAE \cite{Higgins2017betaVAELB} with unit variance corresponds to the case where the first term of the loss function (Eq. \ref{eq:crit}) is weighted by $\beta$ and where $v$ is set to one. Such a model is equivalent to a traditional VAE with $v=\beta$.  Note that the learning rate has also to be divided by $\beta$ because the loss functions
 of the two models are proportional. 
In the $\beta$-VAE case, increasing $\beta$ may lower the quality of the reconstructed samples but may improve disentangling properties (in a completely unsupervised manner). Hence special care is needed to set $v$.
In the following, the two hyperparameters $v$ and $\alpha$ are estimated using cross-validation strategies.

\subsection{DVAE for classification and reconstruction}
\label{eq:proc}

The proposed generative model can be used for classification but it also offers the opportunity to transform a sample from a given class to a sample of another class, by modifying the value of the categorical variables $y$ in the latent representation. The reconstruction of a male mesh (resp. female) as a female mesh (resp. male) is carried out as follows:
first, $\mu$ is computed with Eq. \ref{eq:ss2}  by setting $y$ to the true label. 
Then, the latent representation of the mesh under analysis is simply obtained by concatenating
$y$ and $z$ ($z$ is set to $\mu$). By changing the value of $y$ in the latent representation, we obtain the latent representation of the ``same'' individual but of opposite sex. Finally, the reconstruction can be performed with Eq. \ref{eq:ss4}.

In order to test the consistency of the results, we define also a sex preservation procedure. This is the same procedure as the sex change procedure except that we do not change the value of $y$ in the latent representation.

Note that opposite sex reconstruction and same sex reconstruction both require knowledge of the sex of the mesh under analysis : the true label $y$ is needed to compute both $\mu$ (Eq. \ref{eq:ss2}) and $z$. If the sex of the mesh under analysis is not known, we have to replace the true label by its most likely estimate computed with $q_0$.
Then, for the reconstruction (Eq. \ref{eq:ss4}), we can choose to reconstruct the subject either as a man or as a woman by setting $y$ appropriately. 

\section{Experiments}
\label{sec:res}

Our database consists of 752 CT scans from the University Hospital of Saint-Etienne, France, of which 470 subjects are men, 282 subjects are women. 
For each scan, a hip bone mesh is extracted as explained in section 2.
Each point coordinate is normalized so as to have zero-mean and unit-variance. The means and standard deviations are computed using the training dataset (please see the next paragraphs).
In addition to learning a DVAE, we also learn a vanilla-VAE  whose 
architecture is obtained from the one represented in Fig. \ref{fig1} except that  the label $y$ and the computation of $P_y$ are removed. The usual criterion \cite{Kingma2014} is used for training the VAE. We also learn a classifier (denoted C) whose architecture is obtained 
from the one of Fig. \ref{fig1} by keeping only the layers that are useful for the computation of $P_y$. The binary cross entropy loss is used for training C.

\subsection{Evaluation protocol}

\subsubsection{Hyperparameters setting}
In the VAE case, the variance $v$ is estimated automatically along the training process with the method proposed in \cite{simple}: $v$ is computed for each batch as the MSE loss. Regarding the DVAE, several methods have been tested without success to estimate $v$  automatically. This is why the parameter $v$ as well as the parameter $\alpha$ (Eq. \ref{eq:crit}) are set using cross-validation strategies.
It has been observed that the size of the latent space has a limited influence  on the classification accuracy and on the disentanglement properties for a large range of values of $L$ (for $L$=1 to 64). However, using too small values of $L$ leads to an increase of the reconstruction error. $L$ has been set to 16 in all experiments. For a fair comparison, the size of the latent space of the VAE has been set to $L$+2=18. 
Optimization of the parameters has been achieved by the Adam optimization algorithm with a batch size equal to $16$. During training, all models are trained for 600 epochs. We keep the same learning rate of 0.0006 for the first 200 epochs and then decay the learning rate to 0.0003 for the next 200 epochs. For the last 200 epochs, we set the learning rate to 0.0001. 
\subsubsection{Nested-cross validation strategy}
In order to estimate the ability of the models to handle unseen data and to set the hyperparameters $\alpha$ and $v$ for the DVAE, we  follow the  nested cross-validation strategy.
First, an (outer) stratified $5$-fold cross-validation strategy is used to assess the performance of the models. At each iteration, all folds except one are used as training data (it will be denoted TR) and the remaining one is used as a testing data (TE). 
For the VAE and the classifier C,  a model is learned from the whole set TR and the performance of the model is evaluated on TE.
In order to set the hyperparameters for the DVAE case, an inner $K$-fold cross-validation should be used at each iteration of the outer cross-validation. However, this would require training a very large number of models. To make the problem tractable, we instead randomly divide the training set TR  into a validation set denoted V and a training set T (20\% and 80 \%). Afterwards, several models are learned from T using different values for the hyperparameters: a grid search is used for $\alpha$ and $v$ ($\alpha$ and $\sqrt{v}$ take resp. their  value in $\{0.5,1,2,3,4,5\}$ and  in $\{0.7,1,1.3,1.6,1.9\}$).
Once all models have been learned, the set $V$ is used to select the model that provides the best disentanglement, that is, the one that leads to the highest success rate for the sex change procedure (see sec. \ref{sec:metric}). The optimal values of the hyperparameters are those that have led to the selected model. Then, a last model is learned from the whole set TR using the optimal value of the hyperparameters. Finally, the performance of the model is evaluated on TE.
Note that a score can be computed for each fold. We can then derive an average score and its standard deviation. 

\subsubsection{Evaluation metrics}
\label{sec:metric}

In the (semi)-supervised case, evaluating disentanglement is often achieved by visualising the reconstructions while modifiying the value of a latent variable of interest. In our specific case, this can be easily achieved  since the latent variable of interest $y$ is binary (a hip bone is either associated to a man or to a woman).
Consequently, the model is tested on its ability to perform conditional generation according to the sex label (Sec. \ref{quant} proposes quantitative results while Sec. \ref{sec:resq} presents some visual contents). 
The model is also tested on its ability to classify hip bones and to reconstruct the original data. 

For each fold, we compute four different metrics  to evaluate the performance of the model.
The first metric is   $(i)$ the classification accuracy (CA) obtained with $q_0$ (DVAE) or with classifier C. 
The three other metrics are computed using different reconstructions of the mesh under analysis. In order to distinguish between classification errors and reconstruction/disentanglement errors, the true label is used to compute the latent representation. 
The three other metrics are the following.
$(ii)$ the opposite sex reconstruction success rate (OSRSR): 
we reconstruct a male (resp. female) as a female (resp. male) mesh using the sex change procedure  (Sec. \ref{eq:proc}). This procedure is considered as successful if the transformed mesh is classified as a female (resp. male) one using C.
This rate should be high if the sex label $y$ has been been properly disentangled from the other variables of $z$. 
$(ii)$ the same sex reconstruction success rate (SSRSR): 
we reconstruct a male (resp. female) as a male (resp. female) mesh using the sex preservation procedure (Sec. \ref{eq:proc}).  This procedure is considered as successful if the transformed mesh is classified by classifier C as male (resp. female).  
$(iii)$ The reconstruction error (RE) in millimeters. The reconstruction obtained with the sex preservation procedure is compared with the original mesh. 
Before the comparison, the reconstruction is first transformed in the original space of the data by inverting the different normalization steps. Then, the mean of  the euclidean distances between each associated point is computed leading to a score for a given subject. This score is finally averaged over all subjects of the fold. 

\subsection{Experimental performance analysis}

\subsubsection{Quantitative results}
\label{quant}
Results obtained with the DVAE approach are shown in Tab.\ref{tab1}.
\begin{table}[t]
\caption{Results (mean and standard deviation) obtained with the DVAE approach. CA, OSRSR, SSRSR, and RE stand resp.
for classification accuracy, opposite sex reconstruction success rate, same sex reconstruction success rate, 
and reconstruction error.
}
\centering
\begin{tabular}{|m{1.7cm}<{\centering}|m{1.7cm}<{\centering}|m{1.7cm}<{\centering}|m{1.7cm}<{\centering}|}
\hline
 CA & OSRSR & SSRSR & RE \\
\hline
  $99.59 \pm 0.34\% $ & $99.10\pm 0.92\% $ & $100\% $ &$1.647 mm \pm 0.098 mm $ \\
\hline
\end{tabular}
\label{tab1}
\end{table}

Regarding the classification accuracy, the DVAE classifier  achieves a very high prediction accuracy ($99.59 \pm 0.34\%$). This corresponds to a total of 3 misclassifications out of  752 (one misclassification in 3 folds and zero in 2 folds). 
The independent classifier C achieves similar results since 
only three subjects are  misclassified (these are not the same subjects).
As a comparison, Tab. \ref{tabX} gives sex prediction accuracy for  recent works that are based on the manual positioning of a few landmarks. We cannot claim that the proposed method provides better results since all the methods should be compared on the same database (which unfortunately is not available). However, the proposed method yields state-of-the-art classification results while being free of any manual positioning of landmarks. 
Moreover, the method is data-driven and not guided by expert knowledge. It can be easily adapted for the classification from other bones or bone fragments.

\begin{table}[t]
\centering
\caption{Comparison with previous works on sex determination. Note that previous works need manual estimation of variables (such as lengths, angles or landmark positions) while our approach is fully automatic.}
\begin{tabular}{|l|r|l|r|}
\hline
Method & individuals & variables 
& accuracy\\
\hline
CADOES \cite{CADOES2019}& 256 & 40 (manual) & 97 \% \\
DSP \cite{MURA2005, BRUZ2017} & 2040 & 17 (manual)& $>$ 99 \% \\
Nikita et al. \cite{Nikita2020}& 132 & 3 (manual) & 97 \% \\
Ours & 752 & 5000 (autom.) & $>$ 99 \%\\
\hline

\end{tabular}
\label{comparison}
\label{tabX}
\end{table}

In terms of reconstruction error, the DVAE performs similarly to a vanilla-VAE,  with a mean reconstruction error of 1.728 mm, even if the selected values of $v$ at each fold (DVAE) are always larger than those estimated (for each batch) with the method of \cite{simple} (VAE). The selected values of $v$ in the DVAE case are relatively large because it has been observed that small values of $v$ lead to poor disentanglement properties but, interestingly, an increase in $v$ did not increase reconstruction error.
One could object that the comparison of the reconstruction errors may be considered as biased since the true sex label is used to perform the reconstruction in the DVAE case. 
However, the same result is obtained when using the estimated label:  there are only 3 misclassified cases and 
using the true label or the false one leads to reconstructions that are mostly similar, except in some specific regions.

Finally,  excellent results are obtained for the opposite sex reconstruction success rate, and for the same sex reconstruction success rate. 
The reconstruction as a female (resp. male) mesh of a male (resp. female) mesh is well-classified by C in more than $99 \%$ of the cases (OSRSR).  
Moreover, the reconstruction as a male (resp. female) mesh of a male (resp. female) mesh is always well-classified by C in our experiments (SSRSR).  
Note that the accuracy of the classifier C reaches only $97.17 \pm 1.05 \%$
when classifying  data reconstructed with the vanilla-VAE (instead of $100  \%$ in the DVAE case). 
As previously, the comparison with the VAE approach may be considered as biased since the true label is used for reconstruction in the DVAE case. 
However, we can use a sex preservation procedure that does not use anymore the true label, but instead the label estimated by $q_0$. 
In this case, when classifying the reconstructions obtained by the DVAE (by using $q_0$), the classifier C reaches an accuracy of $99.59 \pm 0.34 \%$, which is exactly the accuracy of $q_0$ (see Tab. \ref{tab1}). Indeed, classifying with C  the reconstruction obtained with the DVAE (by using $q_0$)  provides exactly
the same results than classifying the original mesh with $q_0$:
the classifier $C$ is wrong if $q_0$ is wrong. This clearly shows the consistency of the method.
As an example, if a male mesh is considered as a female one by $q_0$, the DVAE will reconstruct this male mesh as a female one so that 
the classifier $C$ will be also wrong.

\subsubsection{Qualitative results}
\label{sec:resq}
The very high value of the opposite sex reconstruction success rate seems to indicate that
the sex variable $y$ has been properly disentangled
from the other variables $z$.
In order to evaluate more precisely the disentanglement properties of the model, each mesh is compared with its reconstructed mesh (by preserving the sex) or with its reconstructed opposite sex mesh. 
Note that these two reconstructed meshes are those computed in the previous section (the true label $y$ is used to compute $z$). 

\begin{figure}[t]
\begin{center}
\includegraphics[width=0.32\linewidth]{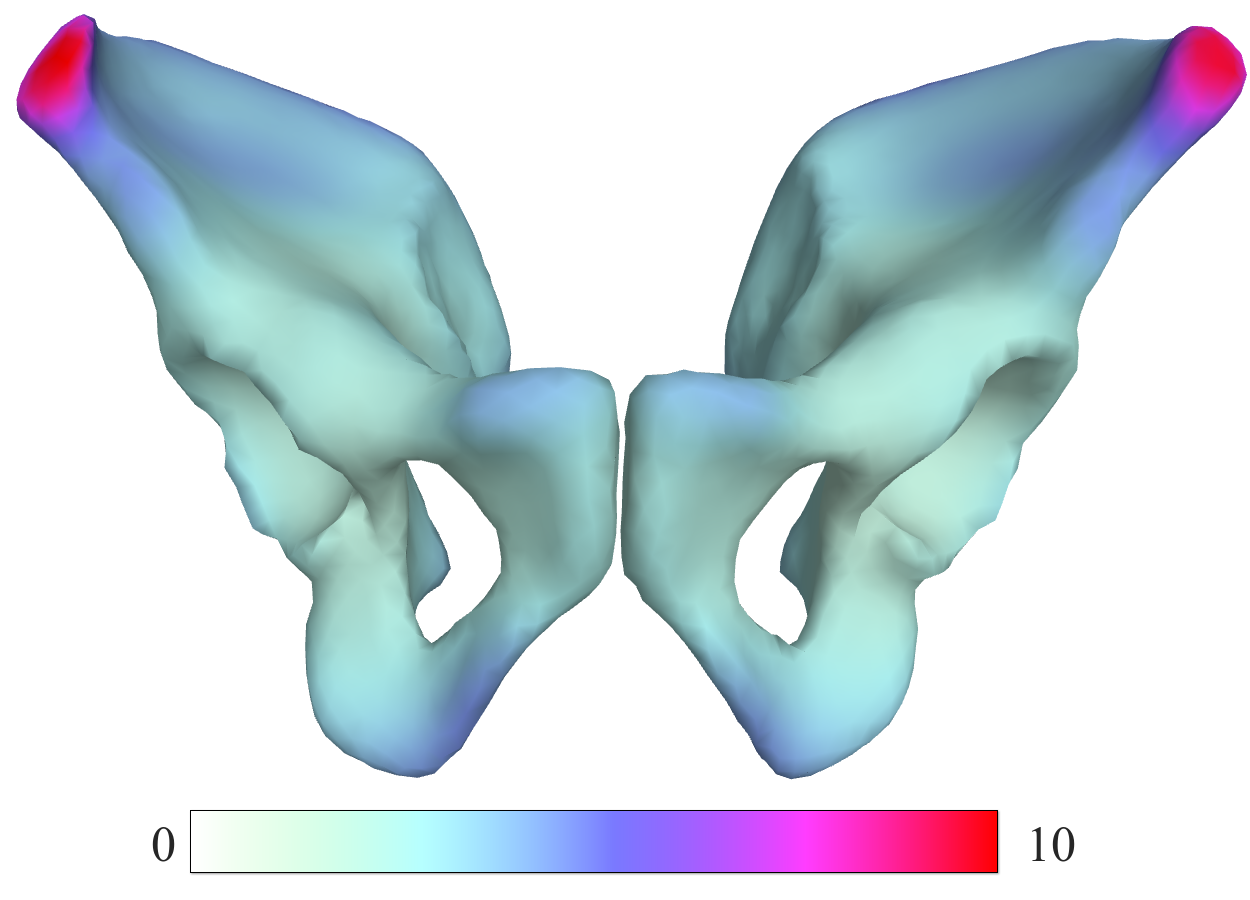}
\includegraphics[width=0.32\linewidth]{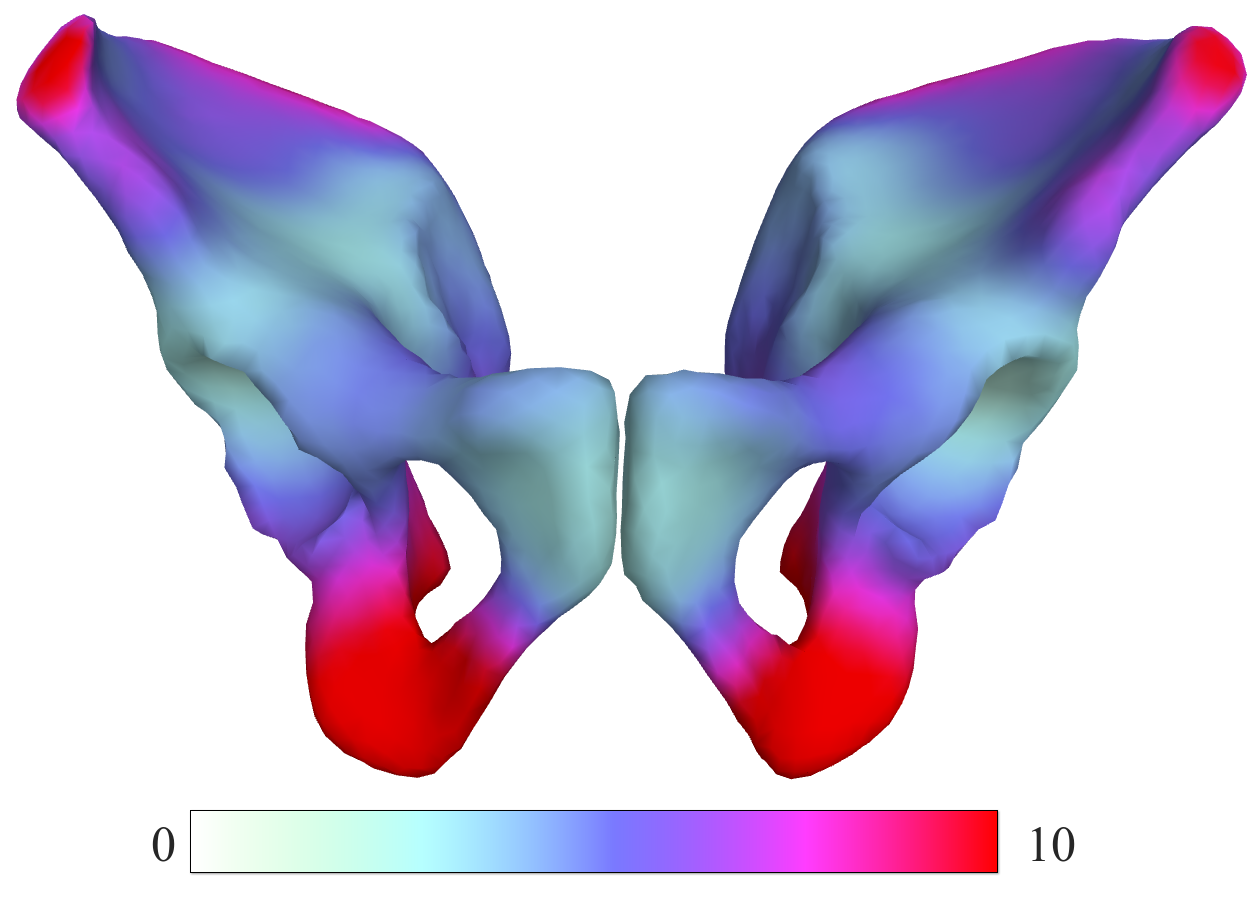}
\includegraphics[width=0.32\linewidth]{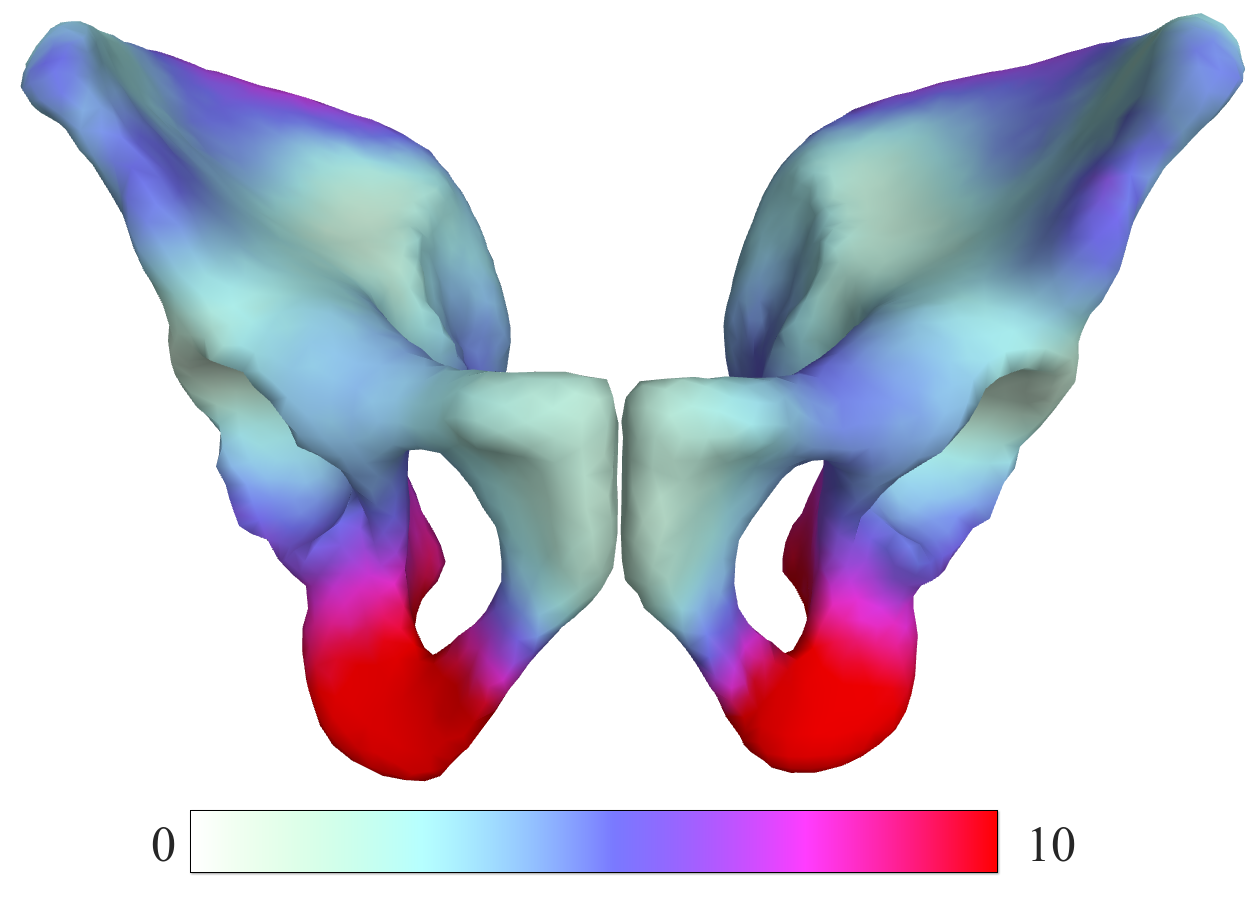}
\end{center}
\caption{Local average distances. From left to right : original meshes vs reconstructed meshes, original meshes vs reconstructed opposite sex meshes, reconstructed meshes vs reconstructed opposite sex meshes. Distances are in mm.}
\label{fig:distances}
\end{figure}

We start here by analyzing averaged results.
Fig. \ref{fig:distances} represents the local distances between the meshes that are averaged across the subjects.
The comparison between the original and the reconstructed meshes (Fig. \ref{fig:distances} (left)) shows that the iliac crest is not very well
reconstructed. This is mainly due to large registration errors that can be observed for some subjects in this region. This makes the problem more difficult because the variability of the data is  increased. 

Note also that opposite sex reconstruction does not alter only a few vertices of the mesh, but changes the  geometry:  
Fig. \ref{fig:distances} represents the local mean differences between the opposite sex
meshes and the original meshes (middle) or the reconstructed ones (right).

The differences that can be observed are consistent with expert knowledge. As an example, the subpubic angle is known to be larger for women, leading to the difference observed in the pubic arch.
Note that the opposite sex reconstructed meshes exhibit differences in the iliac crest with both the original meshes (Fig. \ref{fig:distances}, middle) and the reconstructed meshes (Fig. \ref{fig:distances}, right), suggesting that the iliac crest is a sex-specific region. It is however difficult to give credit to  this conclusion based on the results presented in Fig.\ref{fig:distances} (left): the reconstructed meshes and the original ones exhibit also differences in the iliac crest (the iliac crest is not very well reconstructed). Note that the iliac crest is known to show little sexual dimorphism compared to other areas of the hip bone.
Finally, these results reinforce the idea that the sex variable has been properly disentangled.

\begin{figure}[t!]
\centering
\includegraphics[width=0.4\linewidth]{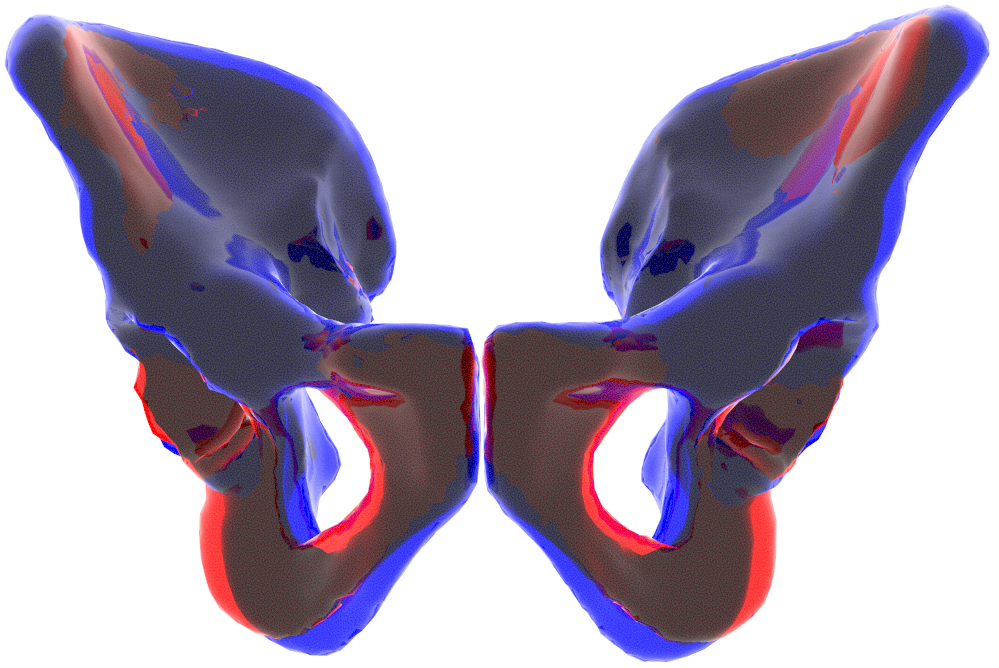}
\hfill
\includegraphics[width=0.2\linewidth]{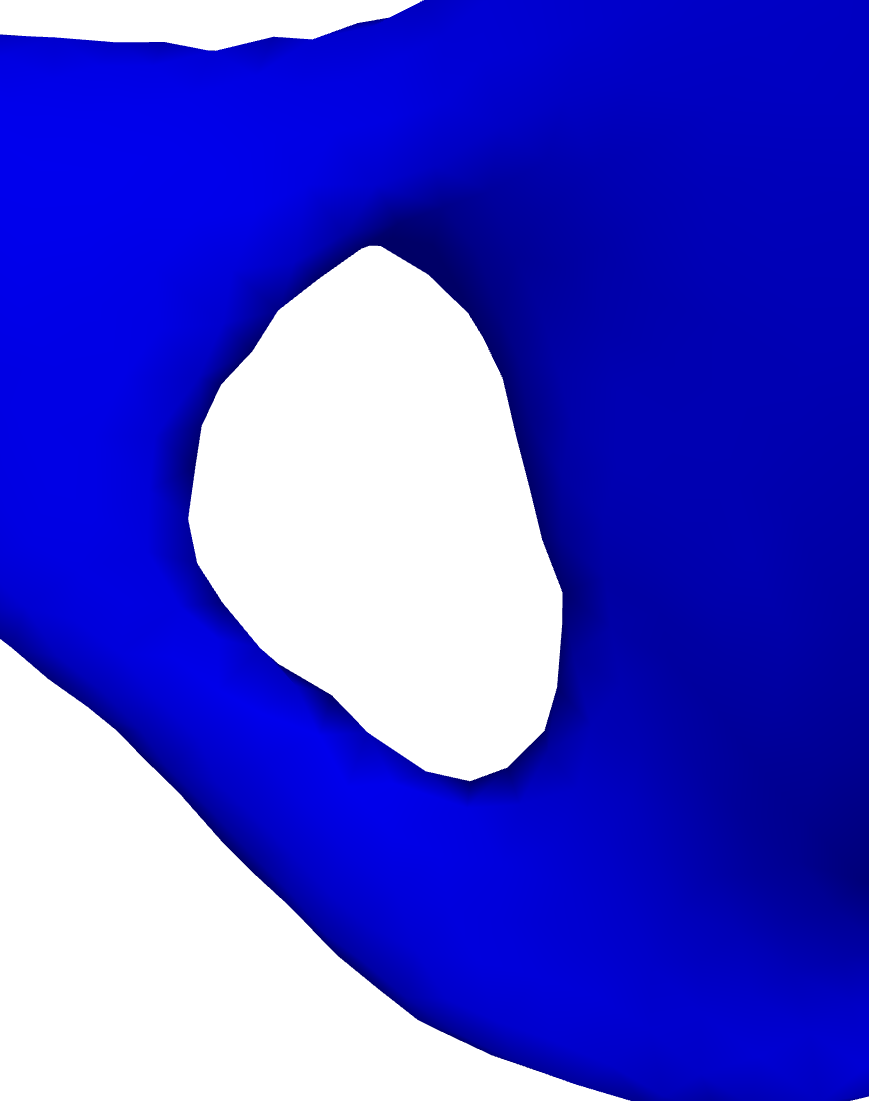} \hfill
\includegraphics[width=0.2\linewidth]{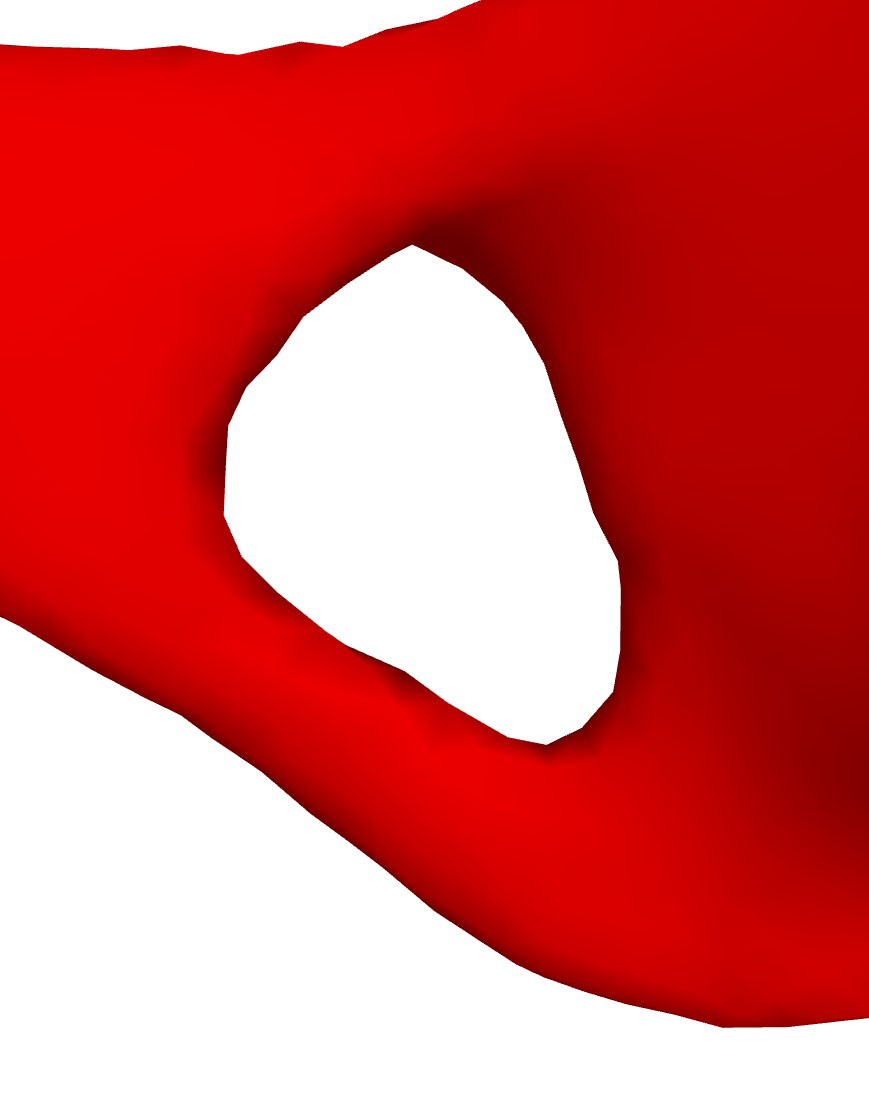}
\caption{Example of changing a male hip bone (blue) to a female hip bone (red). left: angle comparison: the subpubic angle is larger for the female bone than the male bone. right: the male obturator foramen (left) exhibits an oval shape, while the female obturator foramen (right) exhibits a triangular shape.} \label{fig:sexchange3}
\end{figure}

\begin{figure*}[t!]
\centering
\subfigure[]{
\begin{minipage}[t]{0.24\linewidth}
\centering
\includegraphics[height=0.9in]{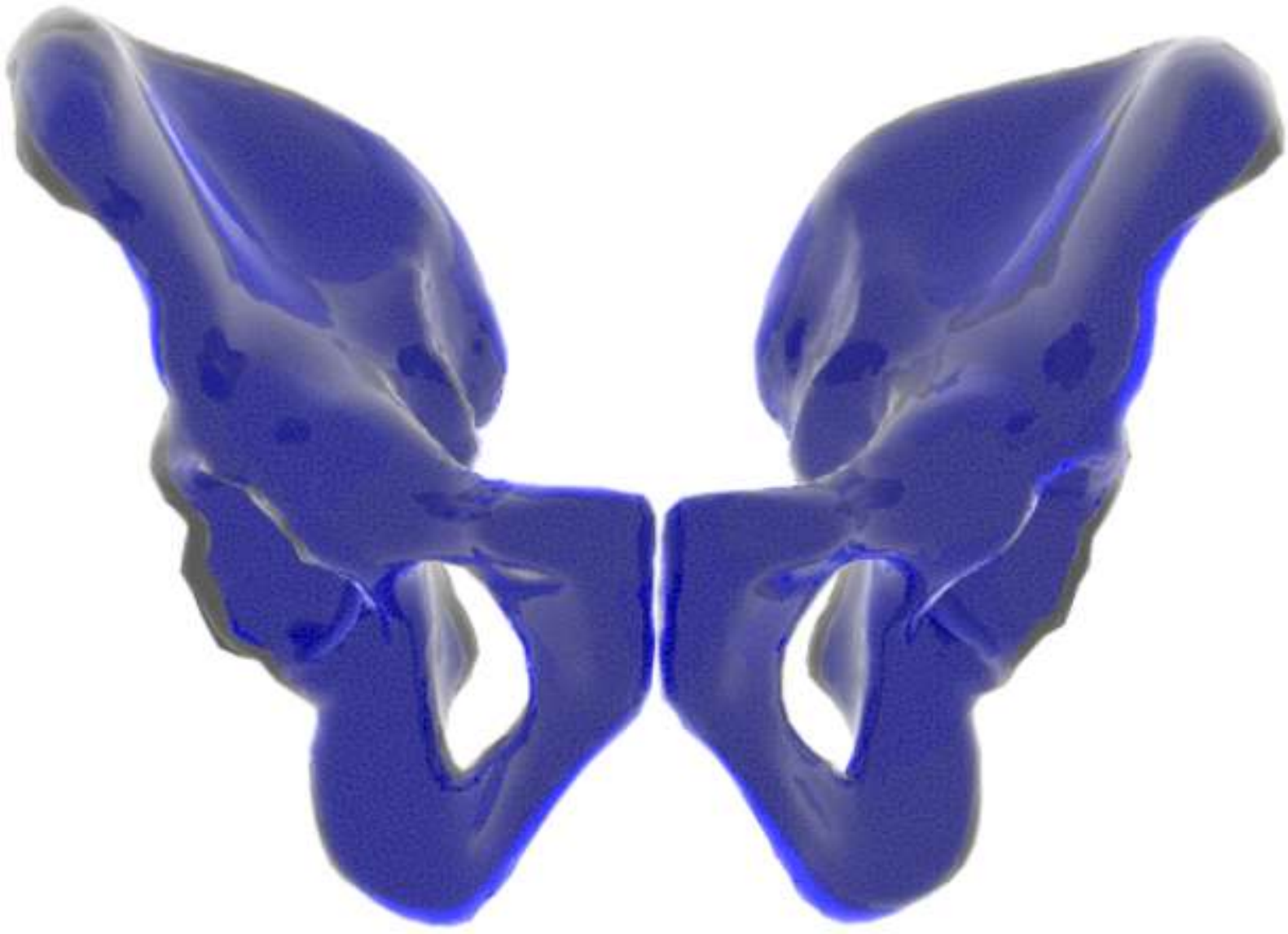}
\includegraphics[height=0.9in]{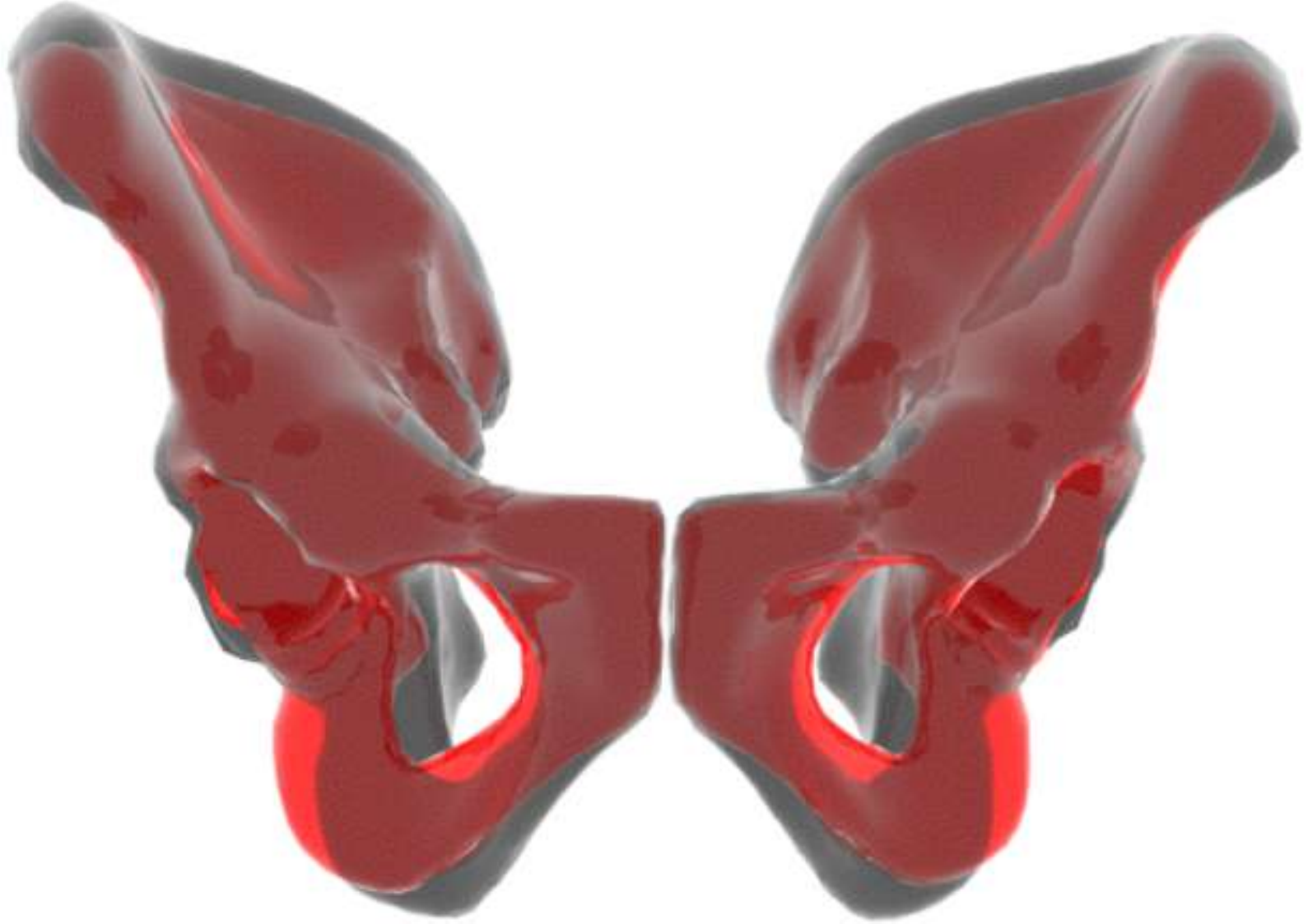}
\includegraphics[height=0.9in]{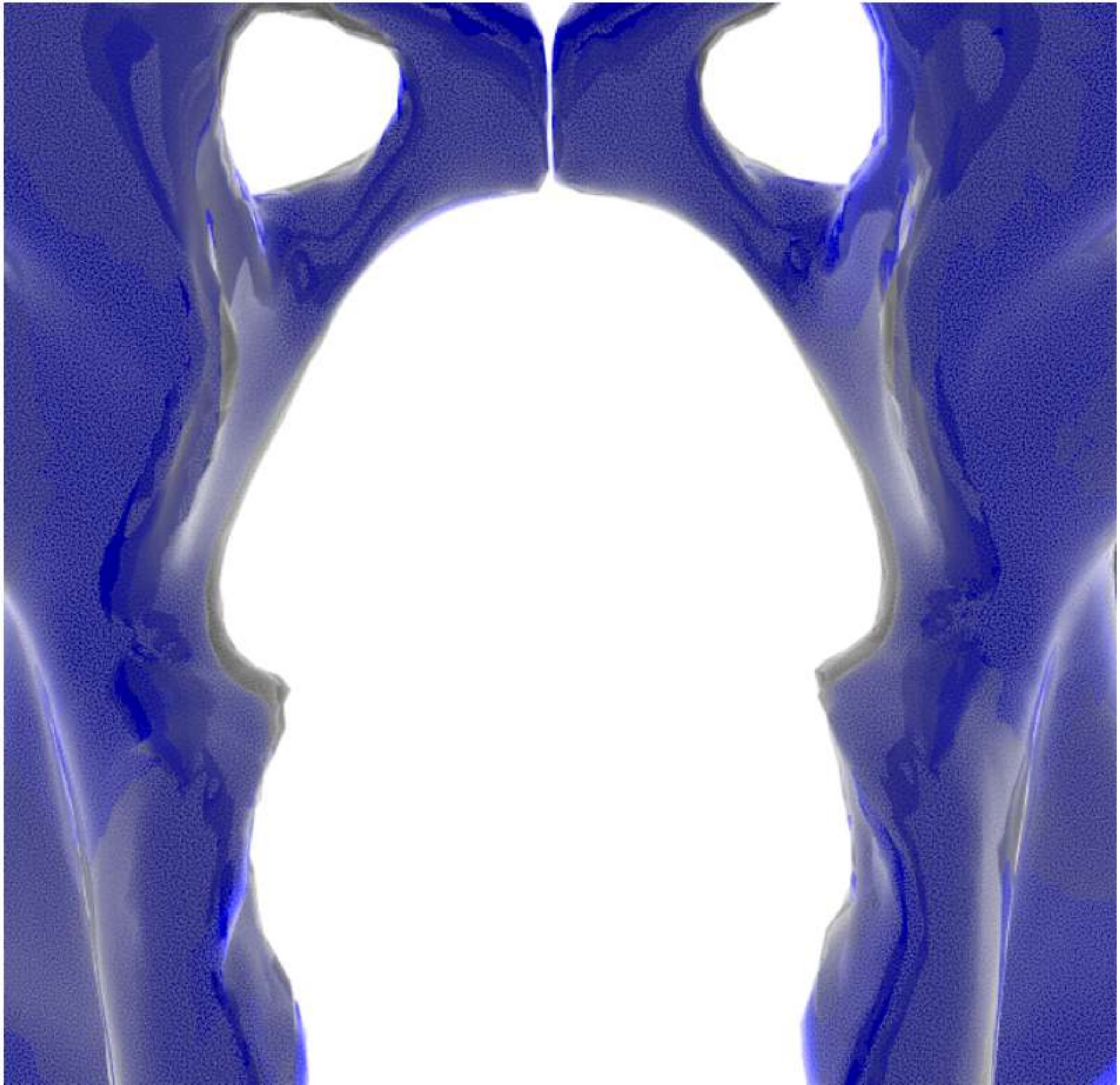}
\includegraphics[height=0.9in]{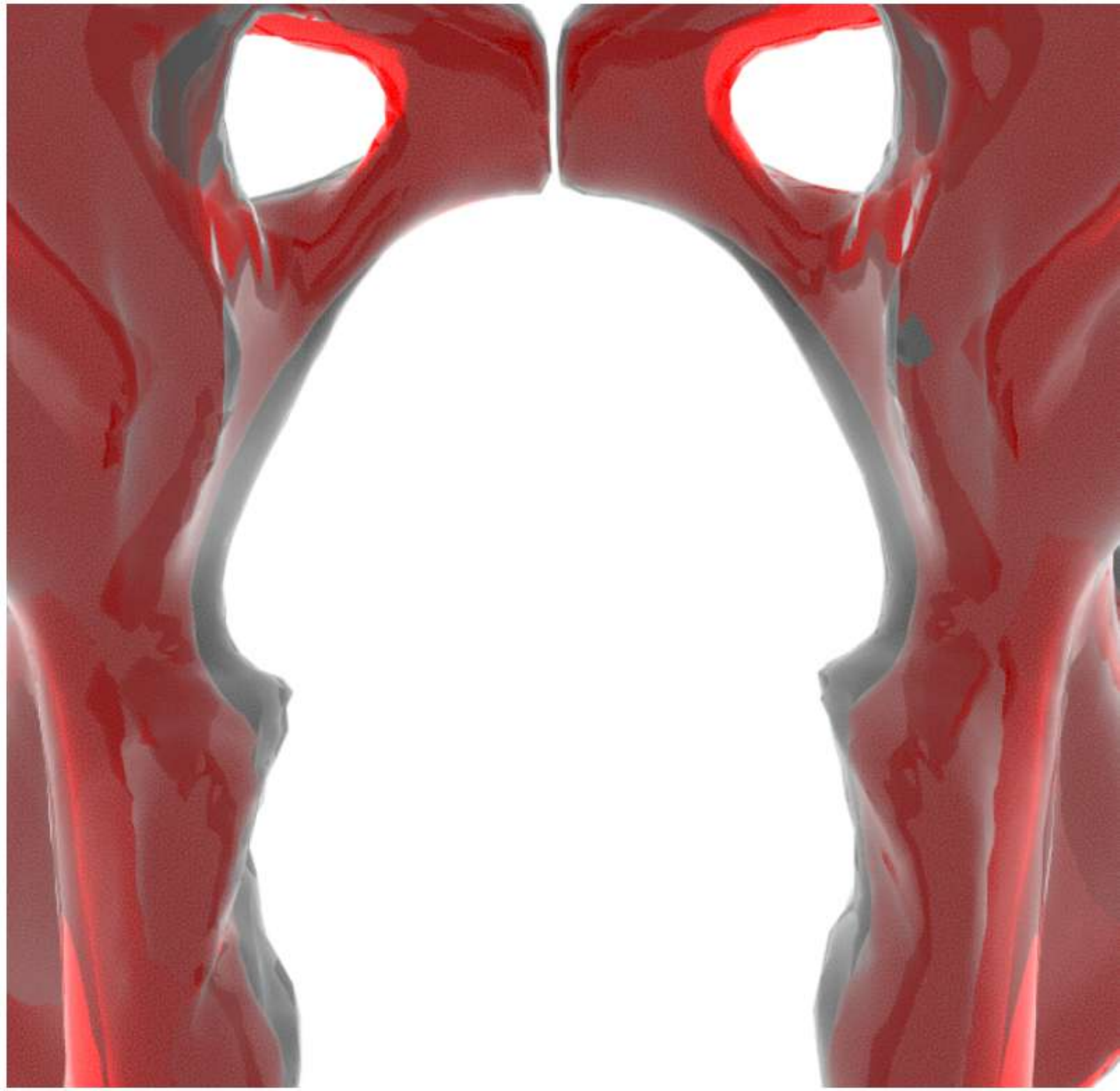}
\end{minipage}%
}%
\subfigure[]{
\begin{minipage}[t]{0.24\linewidth}
\centering
\includegraphics[height=0.9in]{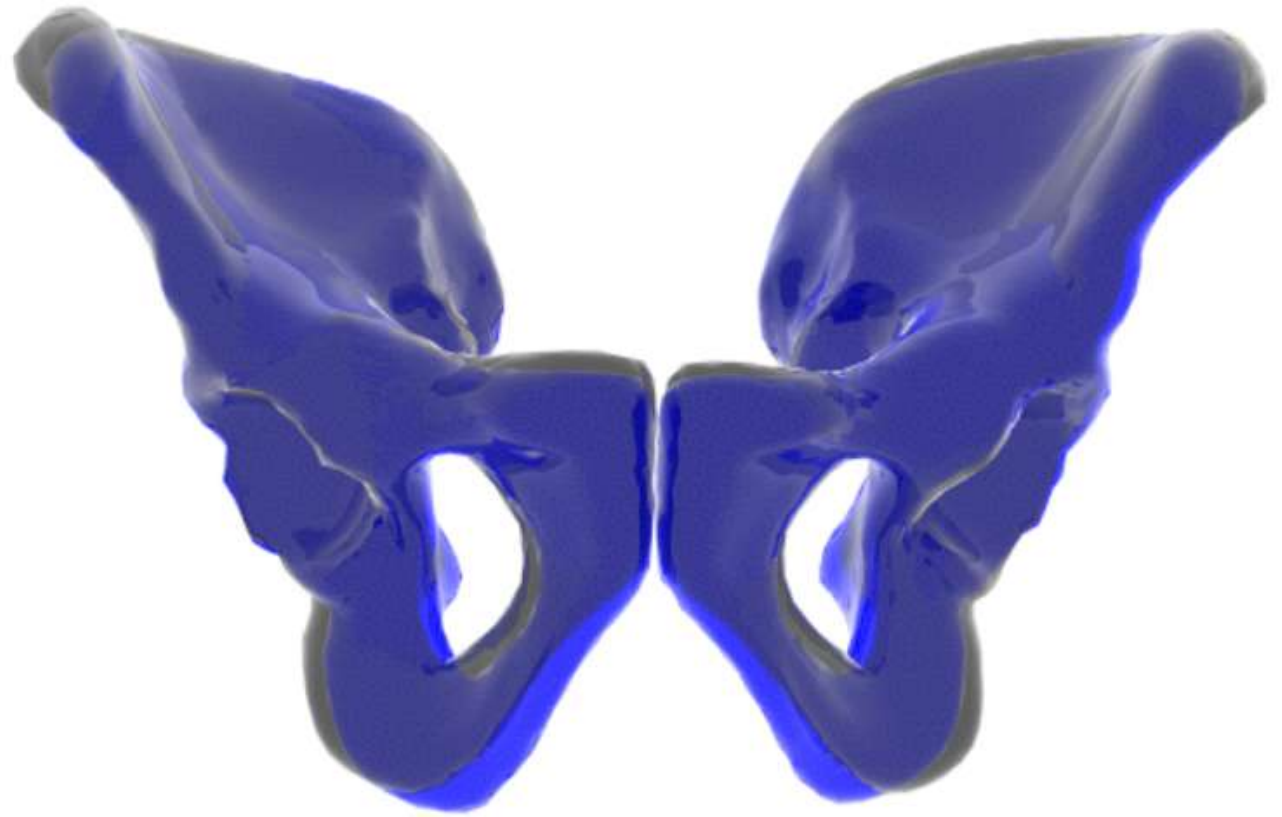}
\includegraphics[height=0.9in]{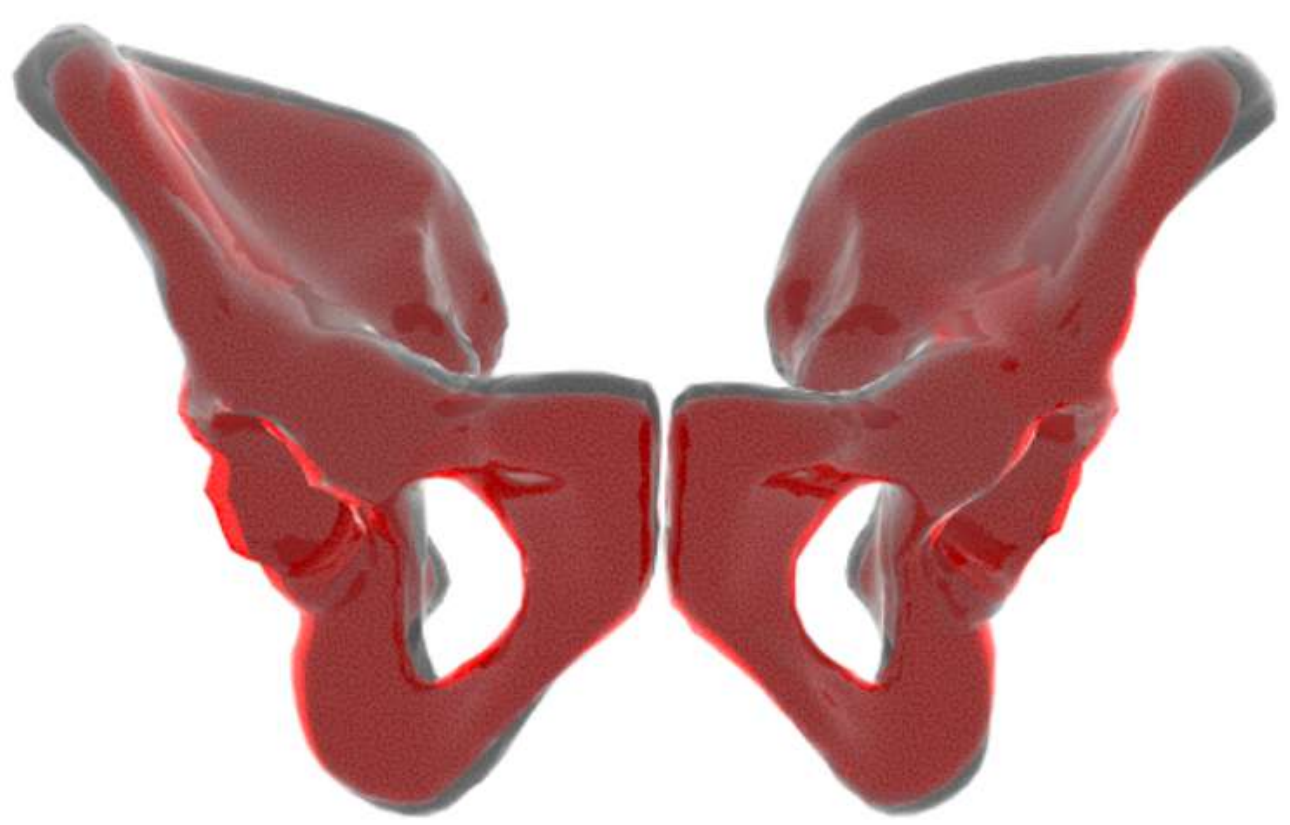}
\includegraphics[height=0.9in]{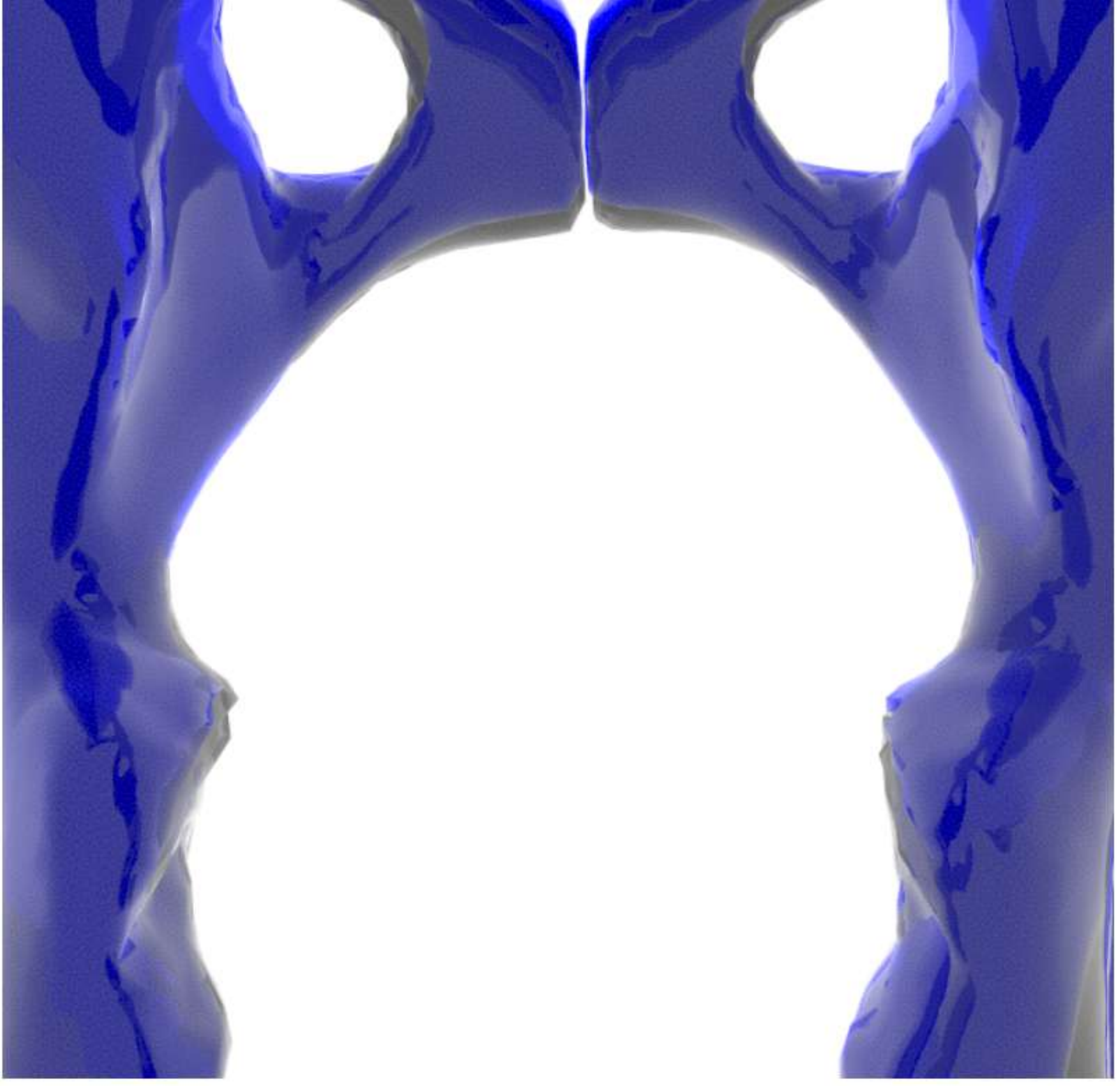}
\includegraphics[height=0.9in]{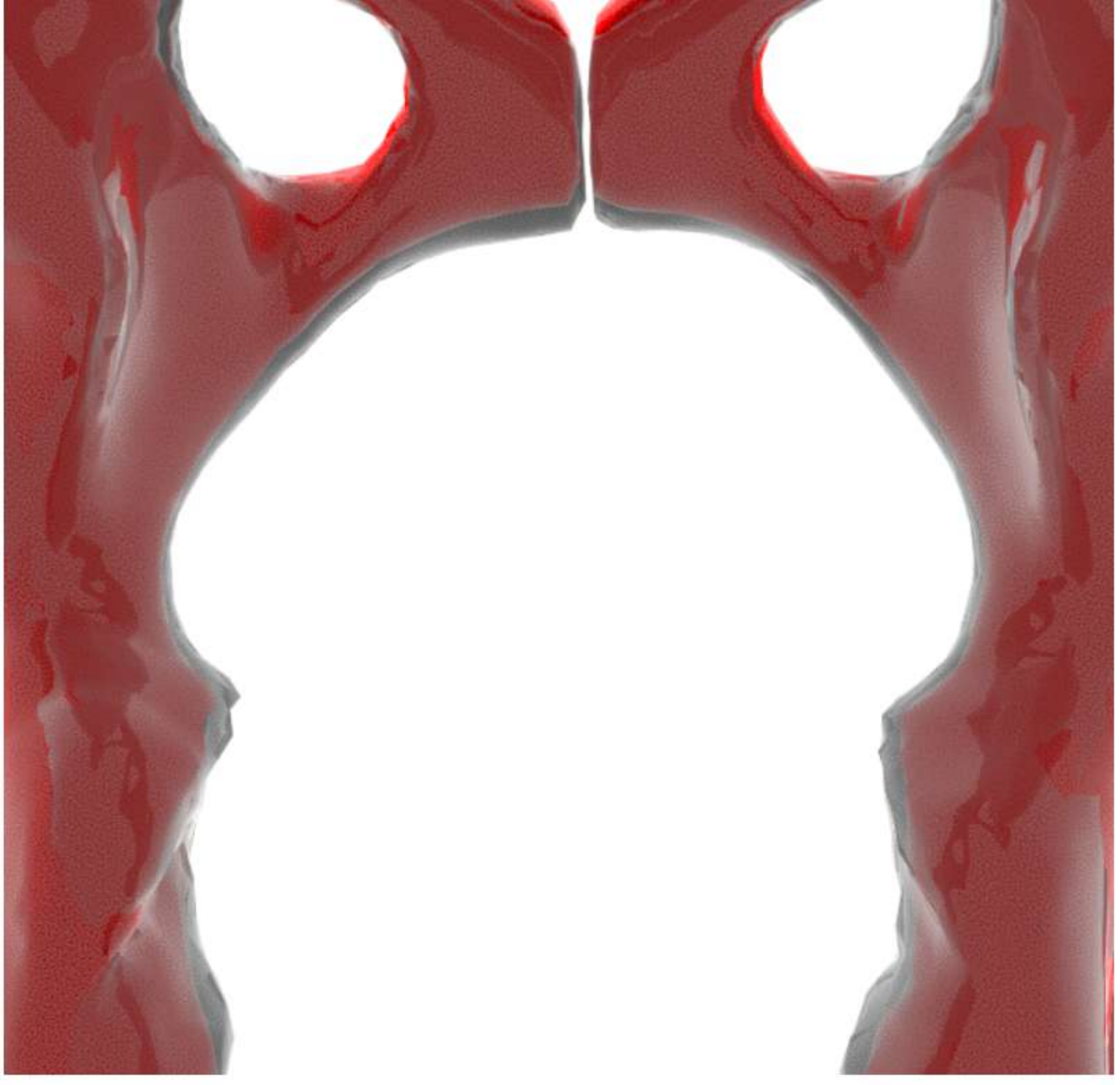}
\end{minipage}%
}%
\subfigure[]{
\begin{minipage}[t]{0.24\linewidth}
\centering
\includegraphics[height=0.9in]{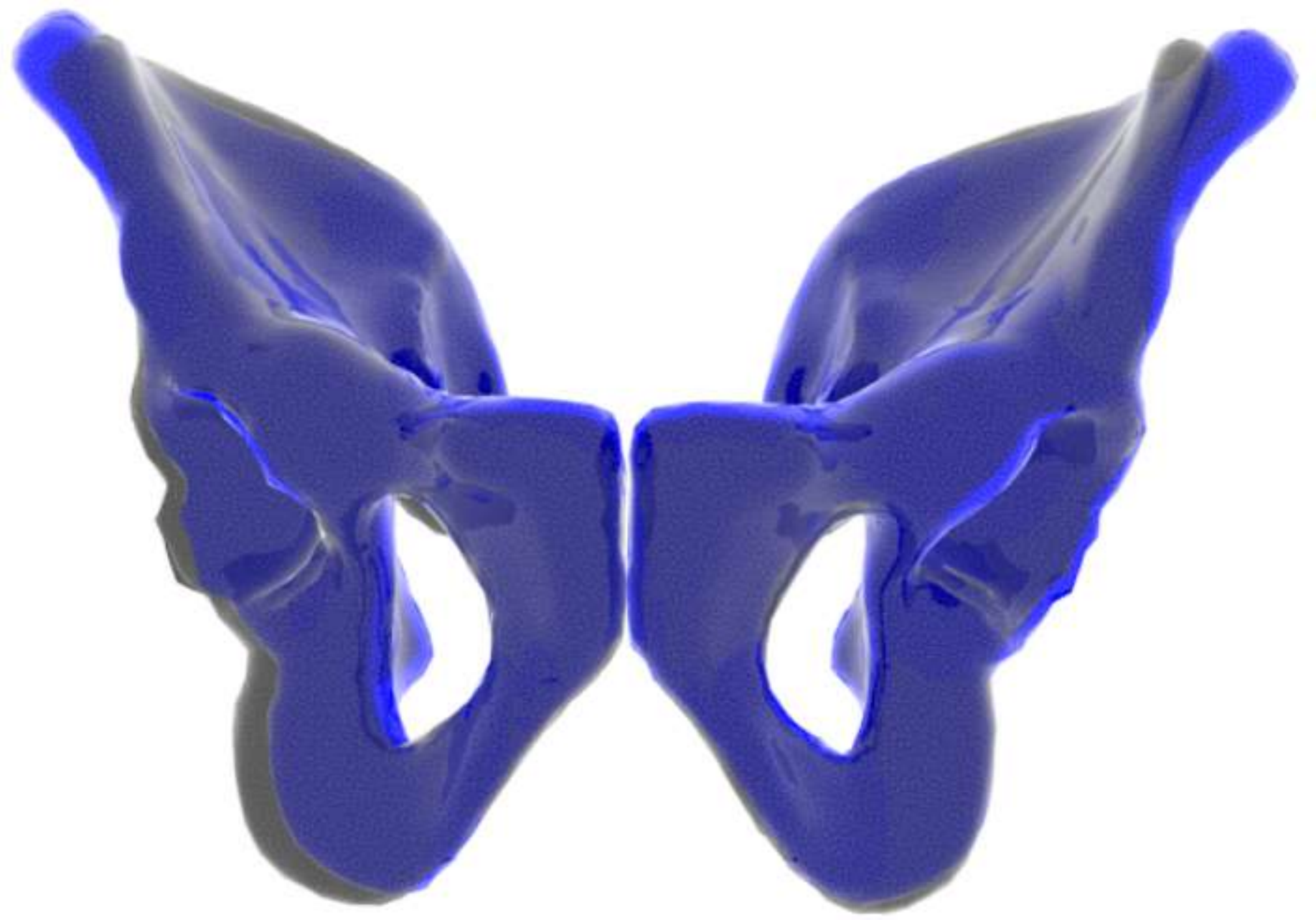}
\includegraphics[height=0.9in]{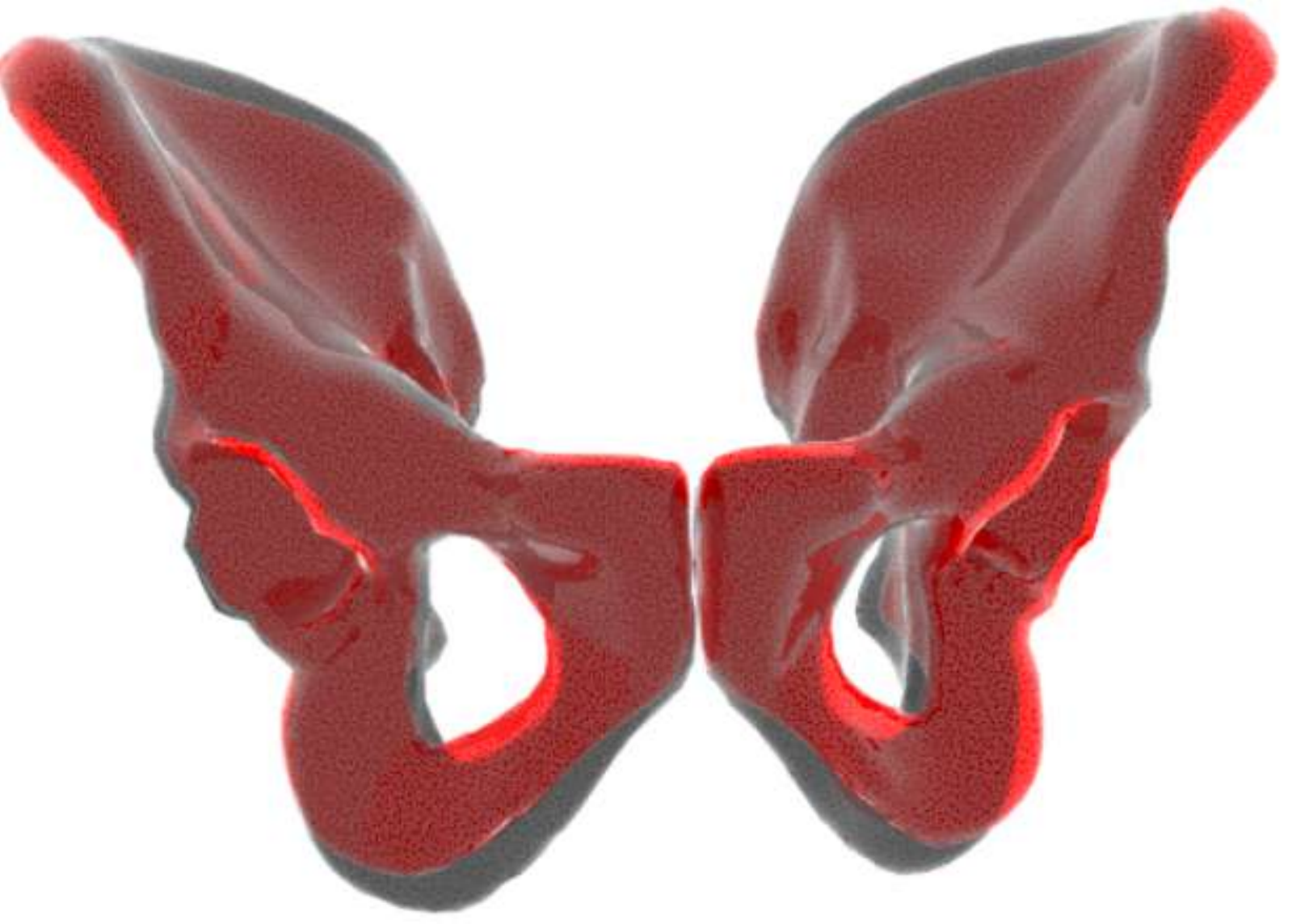}
\includegraphics[height=0.9in]{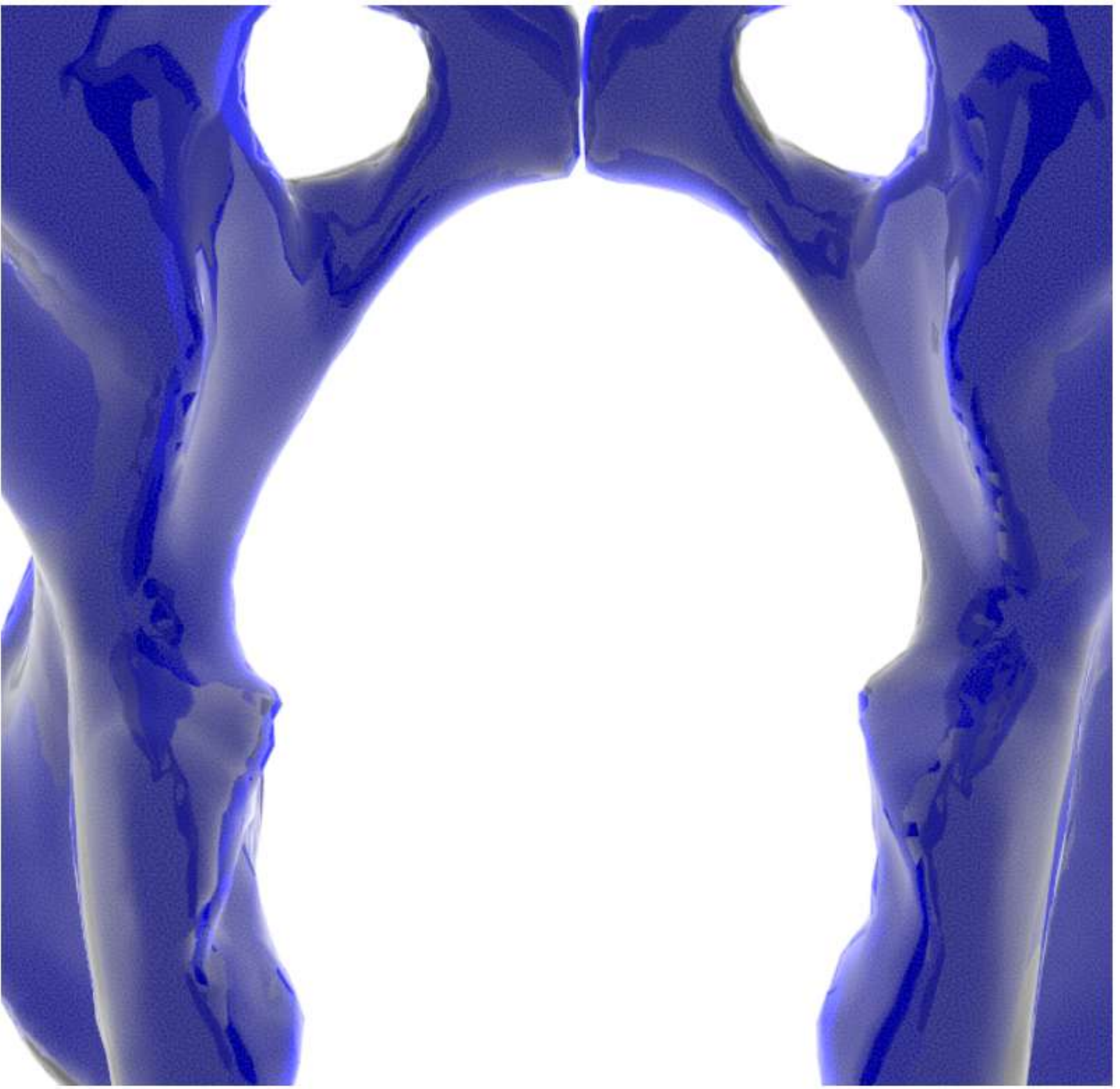}
\includegraphics[height=0.9in]{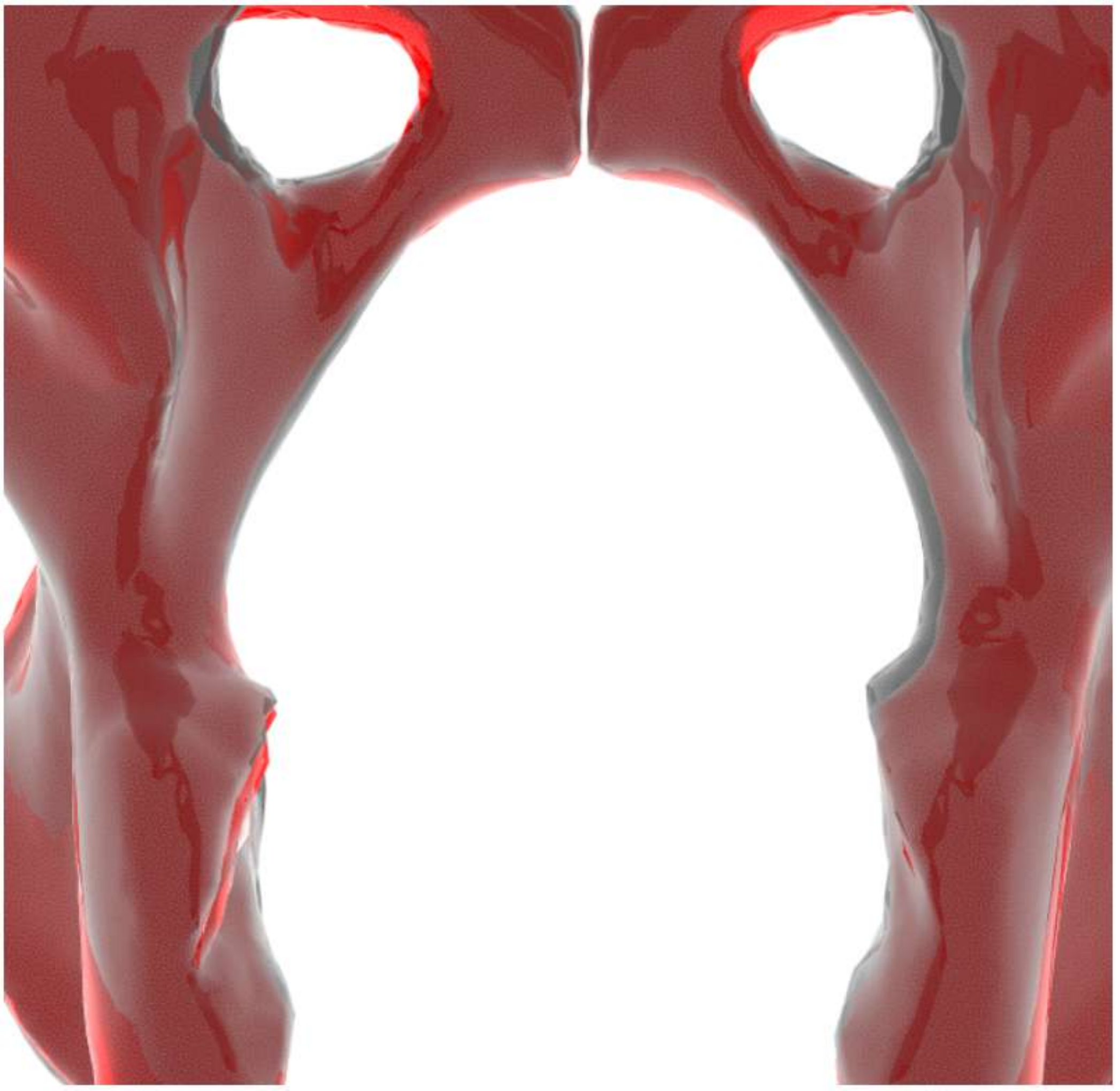}
\end{minipage}%
}%
\subfigure[]{
\begin{minipage}[t]{0.24\linewidth}
\centering
\includegraphics[height=0.9in]{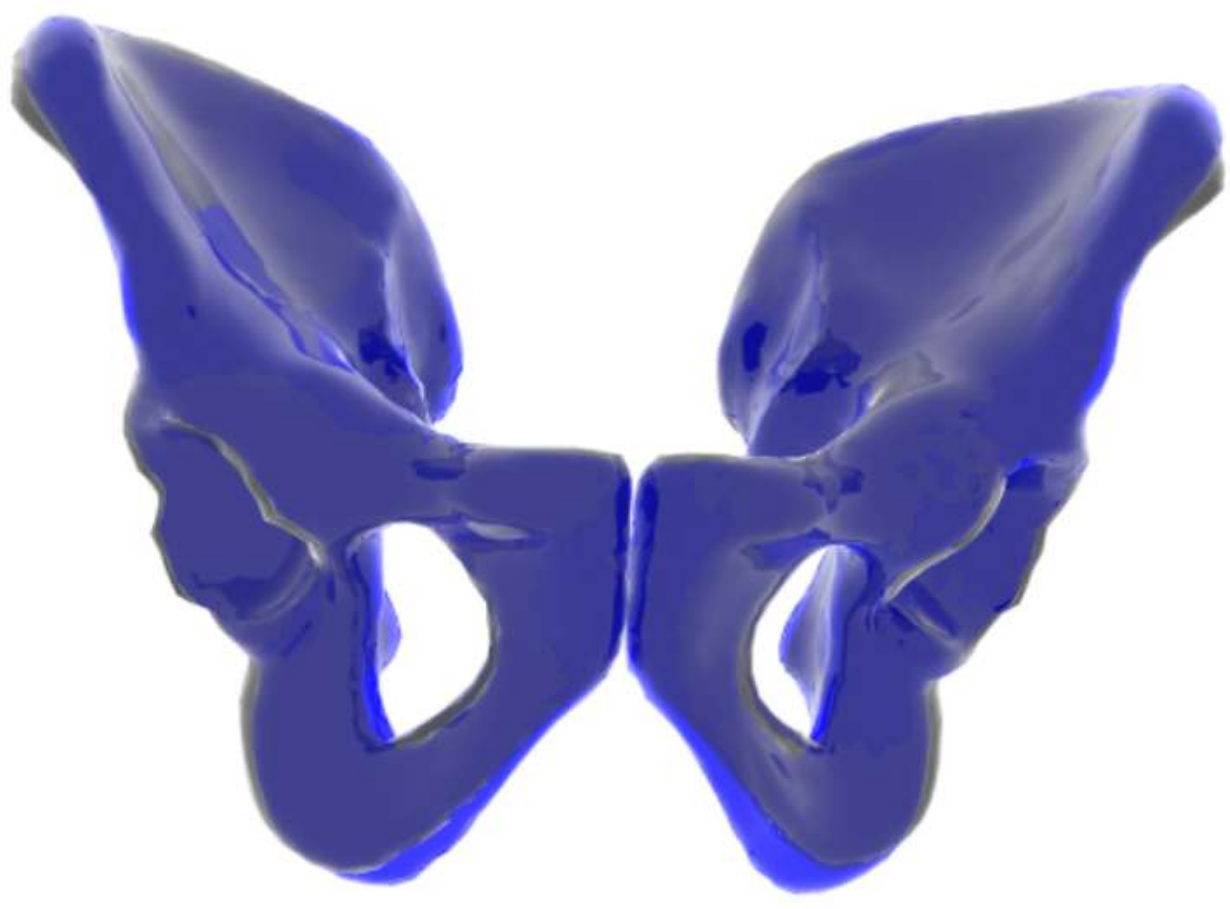}
\includegraphics[height=0.9in]{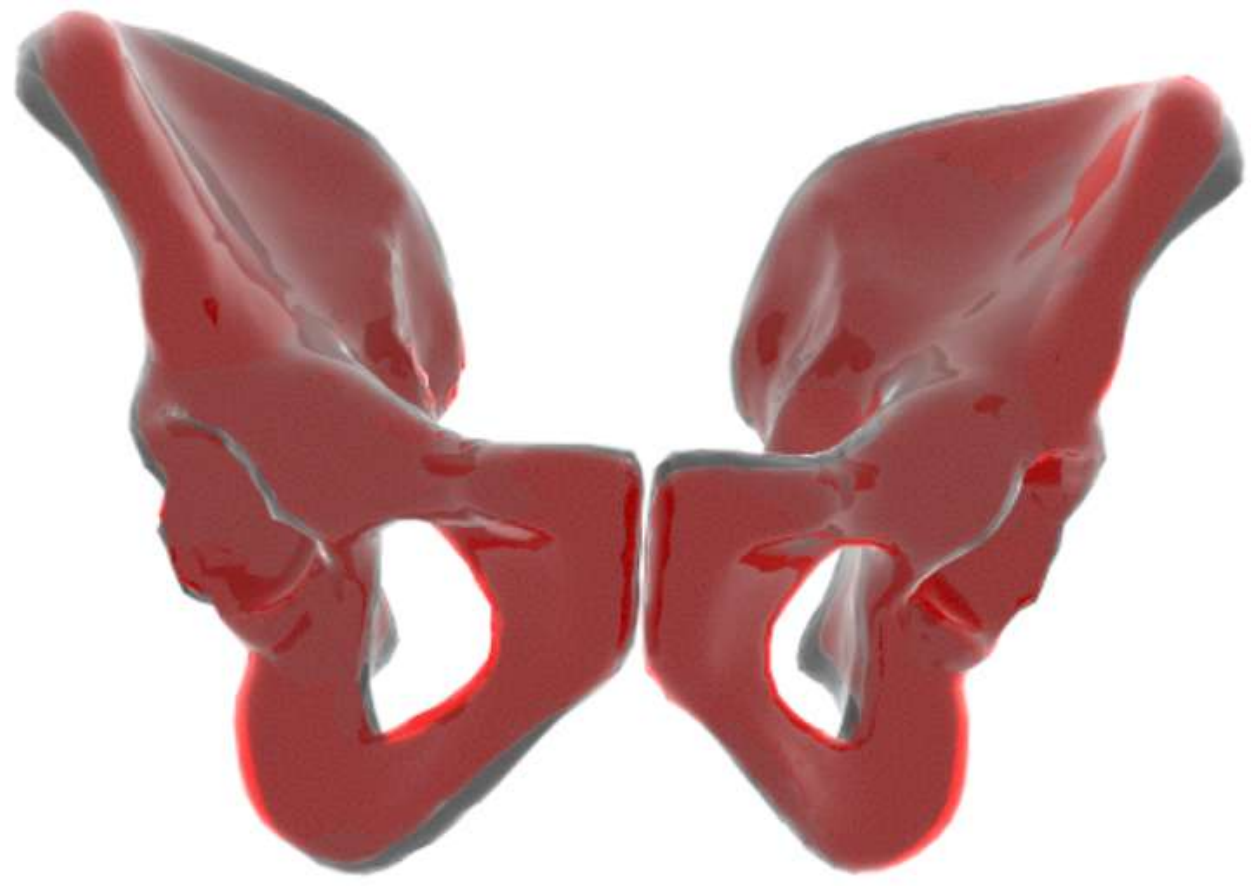}
\includegraphics[height=0.9in]{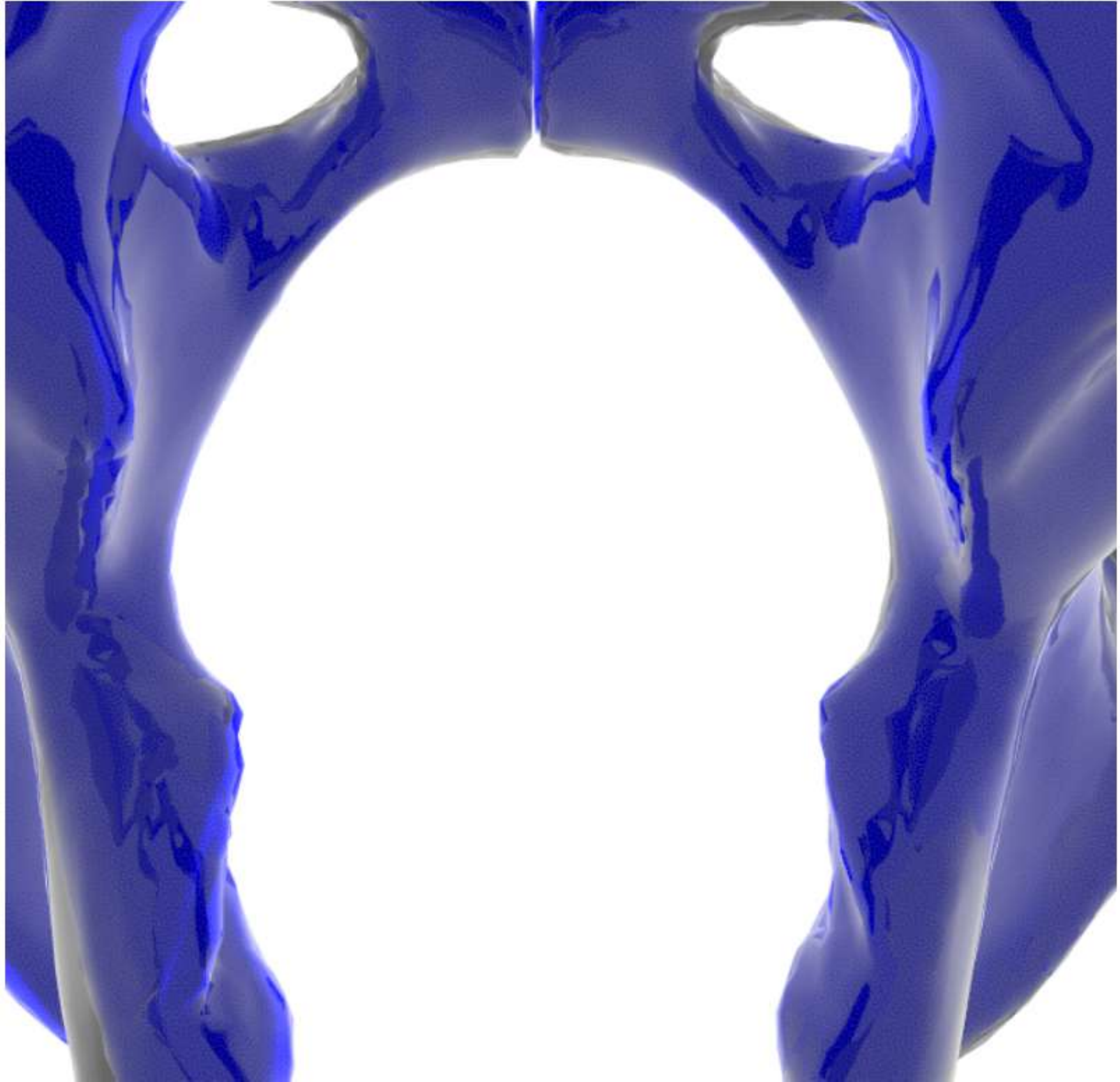}
\includegraphics[height=0.9in]{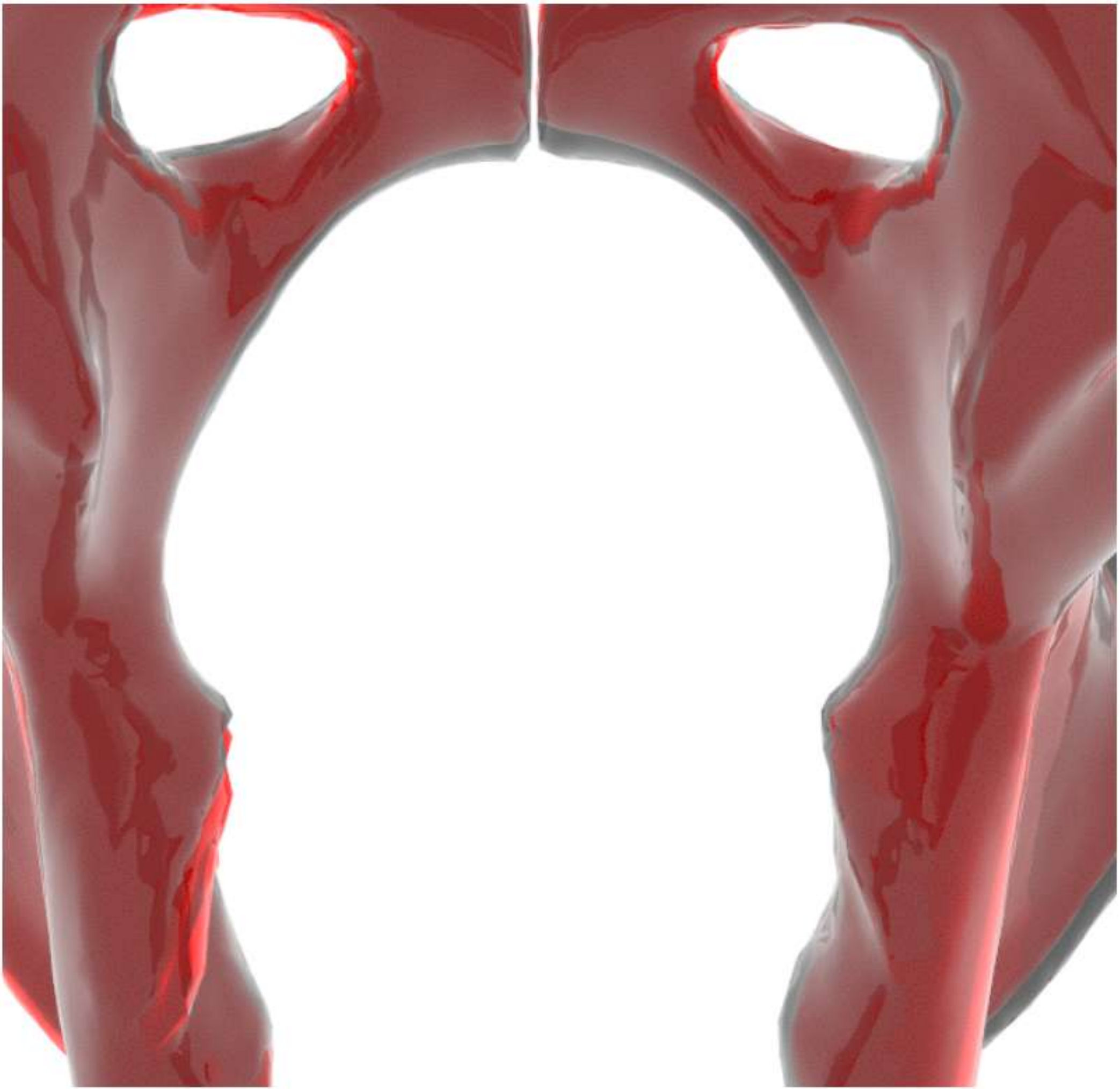}
\end{minipage}%
}%
\caption{Examples of DVAE results. Original mesh (grey) VS mesh reconstructed as a female one (red). Original mesh (grey) VS mesh reconstructed as a male one (blue).
The original mesh of (b) is a female one while those of (a,c,d) are male meshes.}

\label{fig4}
\end{figure*}

We can go further by analyzing individual results. When opposite sex reconstruction is successful, the comparison of  the opposite sex mesh with the original sex mesh (or the reconstructed one) reveals the significant anatomical differences between the male and female hip bones, such as the subpubic angle (Fig. \ref{fig:sexchange3}, left) as well as the shape of the obturator foramen (Fig. \ref{fig:sexchange3}, right), of the greater sciatic notch, of the pelvic inlet and of the symphysis. Note that it may sometimes happen that the two  meshes do not exhibit all the expected differences, but most of them are generally easily observable. 

When opposite sex reconstruction is not successful, the modification is globally consistent, as some significant anatomical differences can be observed, but some of them are sometimes hard to see, or event not present.

\section{Discussion: In what sense does the method provide understanding?}
\label{sec:dis}
Predicting sex from a hip mesh  is not an easy task
for a non-expert and the classification result can be difficult to understand.
In the proposed approach, in addition to providing
the class of the mesh, its reconstructions as a man and as a woman are also provided. When the original mesh is that of a man (resp. woman),  its reconstruction as a man (resp. woman) is very similar to the original mesh. Conversely, the comparison between the original mesh and its reconstruction with opposite sex exhibits differences in some specific areas (while others remain unchanged).
The comparison of these reconstructions with the original mesh enables a
non-expert to make his own choice, or at least also to understand the choice of the classifier. Fig.\ref{fig4}(a) gives an illustrative example of the result provided by the DVAE.
The reconstruction as a man is very similar to the original mesh. 
On the opposite, the reconstruction as a woman exhibits a greater subpic angle 
and a wider pelvic inlet. Consequently, a non-expert can easily classify the mesh as a male  (without using the result of the classifier), or at least, understand why this mesh can be considered as a male one.

It is then legitimate to ask what happens if the label is not correctly estimated: will the proposed method justify a bad result or will it detect the problem?
The different reconstructions relative to the misclassified meshes are shown in Fig. \ref{fig4}(b,c,d) (the estimated label $y$ is now used to compute $z$ so as not to bias the results). 
Since the DVAE has achieved a quasi-perfect classification accuracy, we have only  3 misclassified subjects.
We also analyze in the following the 3 subjects that have been misclassified by classifier C (they are not the same as for the DVAE). 
The 6 misclassified cases can be split into three groups.

The first group is composed of 3 misclassifed subjects (one for C and two for DVAE).
 Fig. \ref{fig4}(c) 
is an illustrative example of this group. It is a man that has been misclassified by C. 
The reconstruction as a man is very similar to the original mesh in the sex-specific regions, whereas the reconstruction as a woman exhibits some differences in these regions. Consequently, the original mesh seems to be a male mesh and it is difficult to understand the choice of the classifier. However, the iliac crest is particularly poorly reconstructed for the considered cases. The shape of this region may be responsible for the misclassifications.

The second group is composed of 2 misclassified subjects (one for C and DVAE).
Fig.\ref{fig4}(b) is an illustrative example of this group. It represents a woman that has been misclassified by DVAE.  
When looking at the subpic angles, everything seems to be consistent and we can assess that this is a female mesh (the reconstruction as a woman is very similar to the original mesh in this area).
However, when looking at the pelvic inlet, everything seems now to indicate that  
this is a male mesh (the reconstruction as a man is very similar to the original mesh in this area). Thus, this mesh has both male and female characteristics. This may explain that this subject is difficult to classify. 
In this case, the two reconstructions enable the user to doubt about the result obtained by the classifier.

The last group is composed of one misclassified subject: 
this is a man (Fig. \ref{fig4}(d)) that has been misclassified by DVAE. 
When it is reconstructed as a woman, the subpubic angle is slightly increased and the pelvic inlet is made wider, as expected.
When it is reconstructed as a man, we expect the reconstruction to be similar to the original mesh but the subpubic angle is slightly decreased. Consequently, the subpubic angle of this man seems to be larger than expected. This may explain why this subject has been misclassified but a user could easily question the result obtained by the classifier, because it seems that the mesh exhibits more male characteristics than female ones.

To conclude, comparing the two reconstructions to the original image is a  simple tool to understand the choice made by the classifier, or eventually to doubt (the second group) about its choice or eventually to question its choice (the first and last groups).
While saliency maps aim to understand what the neural network has learned or how prediction is performed, 
we provide here the user with relevant information so that he can form his own opinion (the proposed approach could be used as a decision support tool). 
Instead of focusing on the neural network, we provide information about the classification problem itself.
Although the two methods provide understanding and explainability, they do not act at the same level. The SM allows to understand the decision process whereas the proposed method allows to reveal the class differences. Finally, these methods are not adapted to the same cases.
When using saliency maps, the user generally knows which information is relevant or not. 
The computation of the saliency  maps is an indirect way of determining whether the neural network can be trusted or not: as an example, it can be questioned if it uses irrelevant information to make its decision. In sharp contrast, we are potentially in the case where the user may need help: it may have difficulties in assessing the sex label from a hip bone (missing data, abnormal morphology), or eventually from a fragment of bone for which it is not known whether its morphology varies or not according to sex. Consequently, our purpose is  to provide the user with relevant information so that he can form his own opinion.

\section{Reconstruction-based classification: application to missing data}
\label{cls}
\subsection{Reconstruction-based classification}

As written in Sec. \ref{sec:dis}, the comparison of the two reconstructed meshes provided by the DVAE approach with
the original mesh enables a non-expert to form an informed opinion.
In the same way, one can wonder if the performances of an independent classifier can be improved by feeding the two reconstructed meshes obtained with the DVAE to the classifier.
To this end, the following paradigm has been used: 
after having trained the DVAE, we train an independent classifier denoted C$_{recon}$ whose
input data are composed of two meshes: the first one is the original mesh from which we subtract its reconstruction as a man (provided by the DVAE, $z$ is computed
using the label estimated by $q_0$)
and the second one is the original mesh from which we subtract its reconstruction as a woman. The classifier C$_{recon}$ is identical to C except the first layer that takes an input of size 4998x6 (we have points in $R^6$ because we model two meshes).
In the following, we denote this method DVAE+C$_{recon}$.

DVAE+C$_{recon}$ achieves an accuracy of $100\%$ for each fold, even with meshes having both female and male characteristics (Sec. \ref{sec:dis}).
One possible reason for these results is that the work of C$_{recon}$ is much simpler than the one of $C$. 
Indeed, let us consider the case of a male mesh. Its reconstruction as a man is very similar
to the original mesh so that the first three components of the mesh (we have points in $R^6$) are close to zero. On the opposite, the reconstruction as a woman exhibits differences in some sex-specific regions so that 
the last three components of the mesh are close to zero except in the sex-specific regions. Consequently, for a male mesh, all components are expected to be close to zero except the last three  components that lie in the sex-specific regions. For a female mesh, all components are expected to be close to zero except the  three first components that lie in the sex-specific regions. 
By highlighting the regions that allow to distinguish male from female hip bone, the input of C$_{recon}$ is much easier to analyze than the original mesh.

\subsection{Application to missing data}
Since all the classifiers C, C$_{recon}$  and the DVAE have already achieved high accuracies, we propose here to make the problem more difficult by introducing missing data : vertices are removed either on the left-hand, right-hand, lower, upper, front or rear side.
The percentage of missing data is expressed in terms of the percentage of the size of the mesh (in the dimension where the data
is removed).
As an example, when removing data on the lower side, the percentage of missing data is expressed in terms of the percentage 
of the height of the mesh.
A very simple imputation strategy is used: missing values are set to the value 0 (which is the mean at each vertex). 

Data augmentation is required during training to achieve good results:  
with a probability of 0.6, the mesh is not modified. Otherwise, it is augmented as follows. 
The side where the vertices are set to 0 is chosen with a uniform distribution, and the percentage of missing data is selected with a uniform distribution in $0-40\%$. 
Four different methods are used for classification.
\begin{itemize}
\item[1] the classifier C.
\item[2] the DVAE: note that the second term of the loss function (Eq. \ref{eq:crit}) uses the original mesh (and not the augmented one) since we want the reconstruction to be similar to the original mesh.
\item[3] the paradigm described in the previous section DVAE+ C$_{recon}$. First, the DVAE is learned as in the second point. 
Then, during the learning of C$_{recon}$, the two reconstructions of an augmented mesh are computed using the DVAE ($z$ is computed using the label estimated by $q_0$) and the input of C$_{recon}$ corresponds to  the augmented mesh from which we subtract its reconstructions. This means that C$_{recon}$ is somehow fed indirectly with augmented meshes during the learning.
\item[4] the last method denoted VAE+C consists in classifying with C the reconstruction provided by the VAE. The VAE is learned in a similar way as the DVAE. Since the VAE provides a reconstruction without any missing data, the classifier $C$ is learned with non-augmented meshes.
\end{itemize}

\begin{figure}[t]
\centering
\includegraphics[width=0.95\linewidth,trim=25 20 43 25,clip]{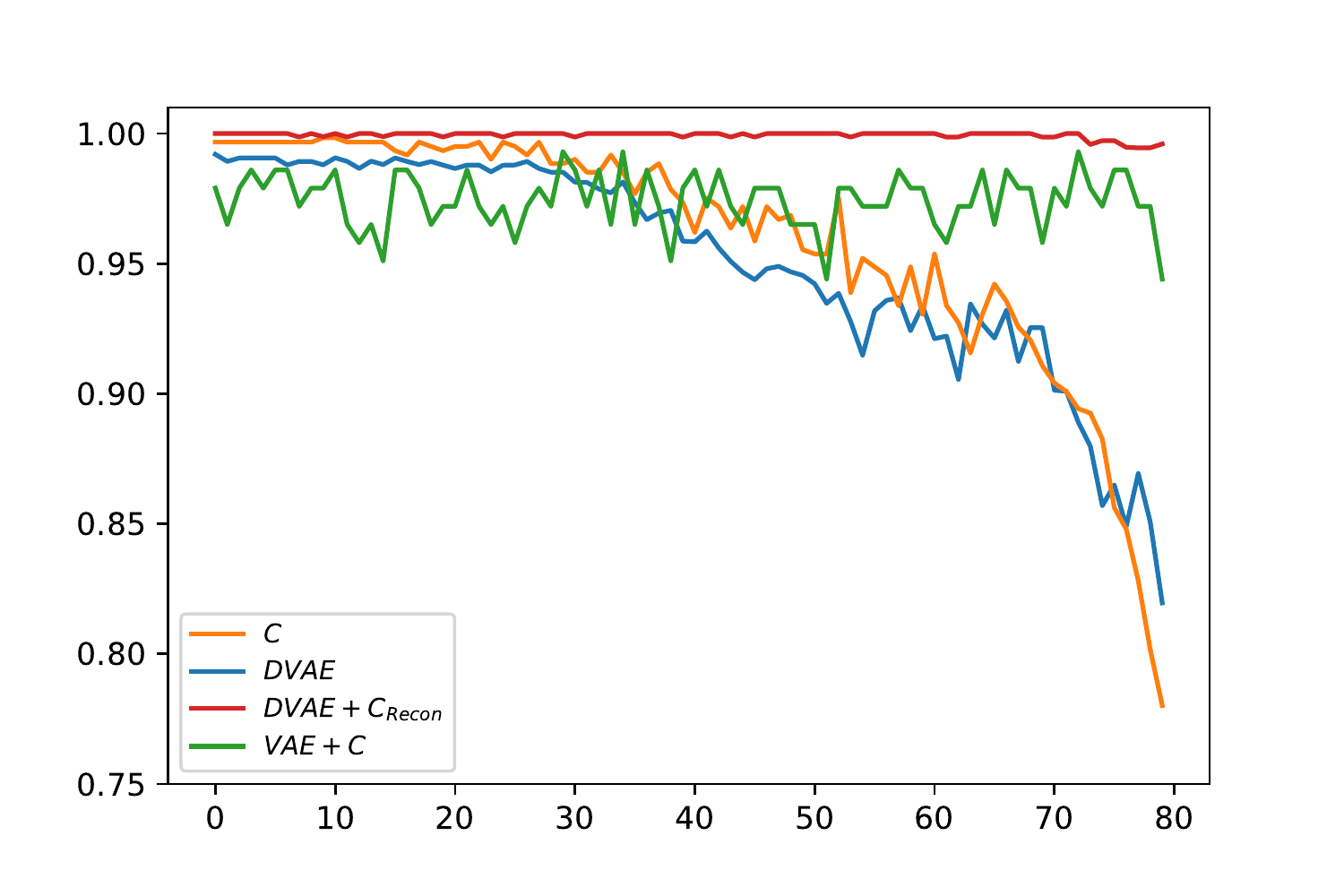}
\caption{Classification accuracies obtained with different methods in the presence of missing data. The x-axis corresponds to the percentage of missing data ($\times$100). }\label{fig5}
\end{figure}

Classification accuracies are shown in Fig. \ref{fig5} for a 
large range of missing data. 
As previously, we can note that DVAE and C provide similar results. Even if $70\%$ of the data is missing, C and DVAE can still achieve an accuracy of $90\%$.
We can also note that the two methods  that use the reconstructions (VAE+C and DVAE+C$_{recon}$) are quite robust to 
missing data but DVAE+C$_{recon}$ performs always better than other classification methods. This clearly highlights the interest of feeding indirectly the classifier with the two reconstructed meshes provided by the DVAE. 

\section{Comparison with saliency maps}
\label{sec:saliency}

\begin{figure}[t]
\centering
\includegraphics[width=.3\linewidth]{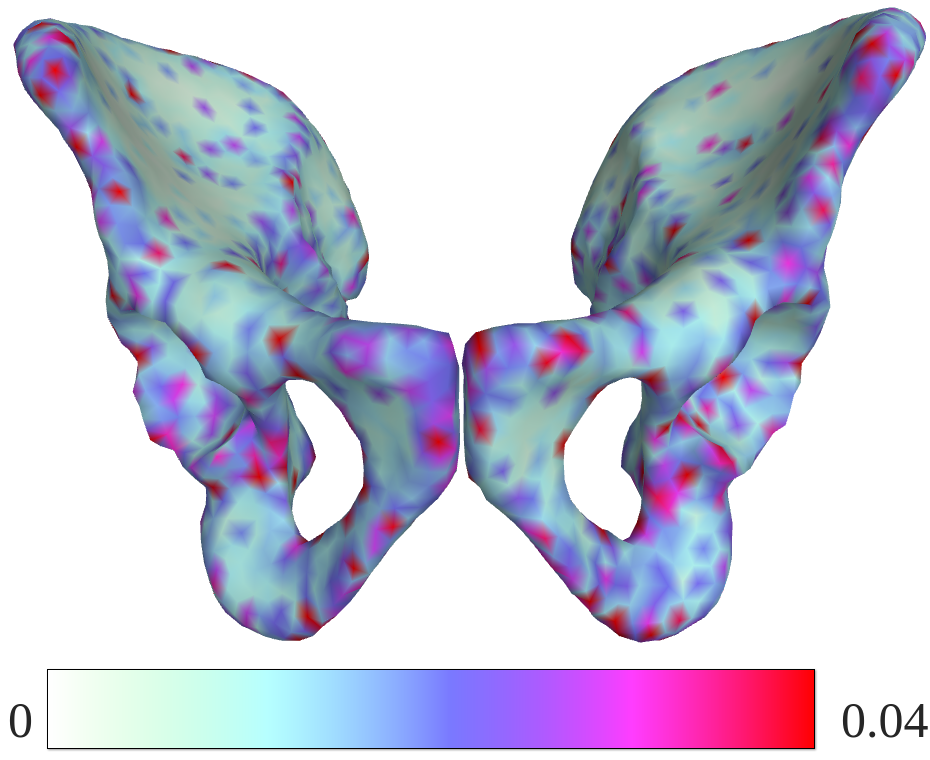}
\hfill
\includegraphics[width=.3\linewidth]{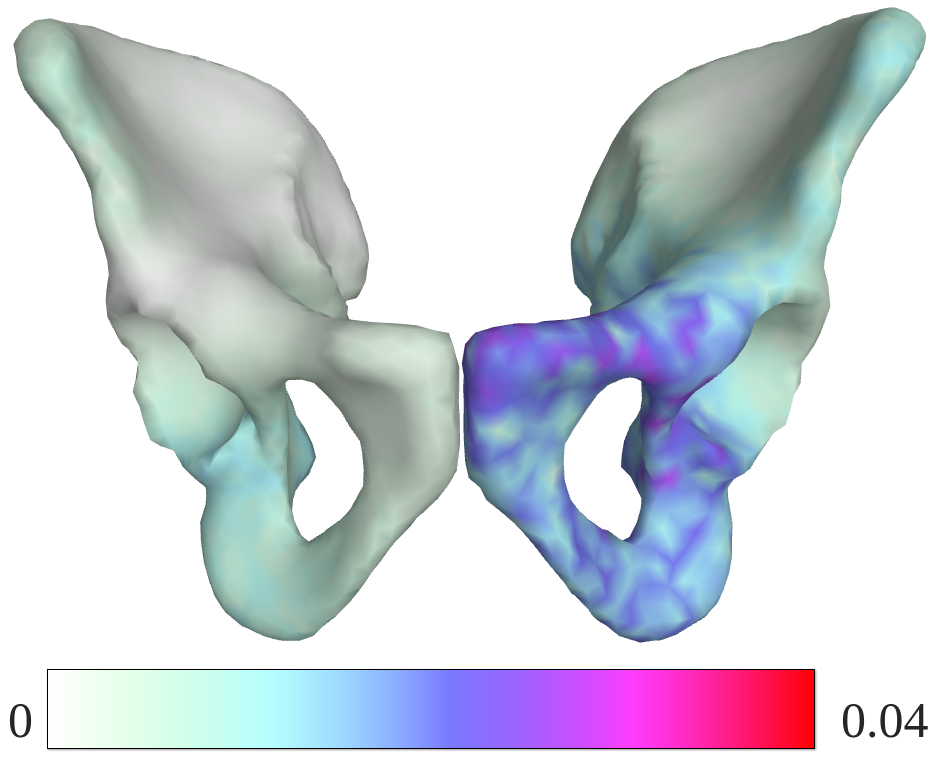}
\hfill
\includegraphics[width=.3\linewidth]{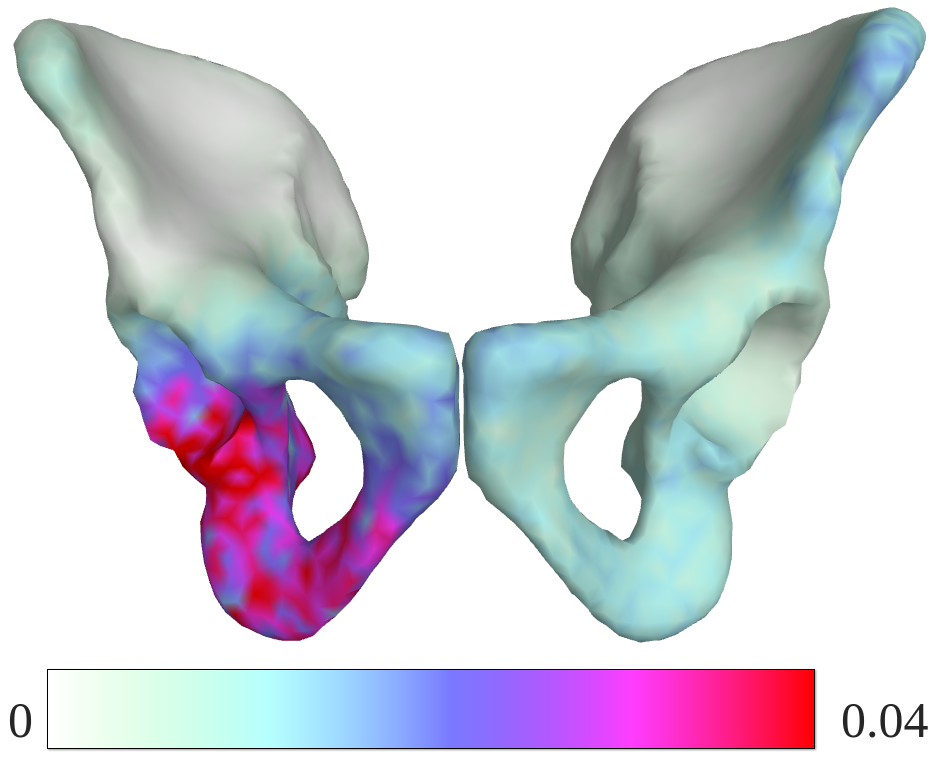}
\caption{Mean saliency maps for C (left) and C$_{recon}$ (center and right). The saliency maps for C$_{recon}$ are either averaged across the female hip bones (center) or the male ones (right).}  \label{fig:saliency}
\end{figure}

To compare our approach for the interpretation of mesh classification with a standard method, we have computed saliency maps for the classifiers C and C$_{recon}$ (without missing data) with the method in \cite{simonyan2014deep}.
For a given input mesh, the importance $w_{ic}$ at each vertex $v_i$ is computed as follows: 
\begin{equation}
w_{ic}=|\frac{\partial p(y=0|x)}{\partial x_{ic}}| = |\frac{\partial p(y=1|x)}{\partial x_{ic}}|, 
\label{equ:saliency}
\end{equation}
where $x_{ic}$ ($c=$1, 2 or 3) represents either the $x$, $y$ or $z$ coordinate.
Eq. \ref{equ:saliency} can be computed using back-propagation. For each vertex, the 3 computed importances (one for each coordinate $c$) are aggregated using the max function: the saliency map at vertex $i$ is computed as $max_{c}(w_{ic})$. Instead of considering the derivative of $p(y|x)$, it is also possible to use the unnormalised score (the softmax layer is not considered for the computation of the derivative). In this case, Eq. \ref{equ:saliency} no longer holds and a saliency map is obtained for each class. Regardless of the methods used or the aggregation function used, the results were always very similar. Fig. \ref{fig:saliency} represents the mean of the saliency maps (across the subjects),  computed with Eq. \ref{equ:saliency} and the max aggregation function.

For the classifier C (Fig. \ref{fig:saliency}, left), it is hard to understand how the classifier makes its decision, as the most important vertices for the classification are distributed over the entire hip bone (we could expect them to lie specifically in regions that are known to differ between men and women, but this is not the case).

In addition, it has been observed that the individual saliency maps were very different from one another whereas one would expect that all of them highlight the sex-specific regions. Finally, the results were neither intra-architecture repeatable nor inter-architecture repeatable. 

Our hypothesis is that saliency maps are not suited to classification problems 
where the features that allow to distinguish between the different classes are spatially correlated and scattered. 
Under these conditions, two classifiers can achieve high accuracy results without having the same decision boundaries, hence their respective saliency maps will be different.

For C$_{recon}$,  the map is more consistent with our expectations (Fig. \ref{fig:saliency}, center and right) except that a strong asymmetry is observed depending on whether the processed hip bone is a female one or a male one. That is why, the saliency maps are either averaged across the female hip bones (Fig. \ref{fig:saliency}, center) or the male ones (Fig. \ref{fig:saliency}, right).
Moreover, contrary to the local average distances (Fig. \ref{fig:distances}),  the mean saliency maps highlight the pubic left tubercle, whose shape is known to vary slightly according to the sex (this is clearly visible for the saliency maps associated to women, a little less for those associated  to men). 
It seems that the classifier focuses here on a subtle difference between female and male hip bones. Since the input of C$_{recon}$ is partly fed with the output of the DVAE, it can be estimated that this small difference has been captured by the DVAE. 
Note finally that similar mean saliency maps can be obtained when measuring intra-architecture repeatability and inter-architecture reproductibility. In all cases, the saliency maps associated to men and women highlight a different side of the hip bone except that 
the asymmetry can be more or less pronounced. Moreover, the side of the regions of interest may be permuted: the mean saliency map of male hip bones highlights the regions that are on the right side (Fig. \ref{fig:saliency}, right) but it can be the left side for other tests. Finally, the modified saliency maps obtained with C$_{recon}$ are more satisfactory than those obtained with C. 
Our interpretation is that the input of C$_{recon}$ is much simpler to analyze since the sex-specific regions
have been highlighted by the DVAE: all components that lie in regions that are not sex-specific are close to 0.

In conclusion, the standard saliency map does not provide a semantic meaning related to the highlighted regions. In contrast, thanks to the conditional generation according to the sex, the proposed method reveals the differences between the male and female hip bones: as an example, we clearly observe with the proposed method that the subpic angle is larger for women (Fig.\ref{fig:sexchange3}).

\section{Conclusion}
\label{sec:con}
This paper has presented a new paradigm for the interpretation of classification by neural networks, based on Disentangled VAE representations. The approach has been illustrated on the interpretation of sex determination from meshed hip bones, on a large database of real data. The proposed paradigm is comprehensive and suited to the disentanglement and classification of other factors of general interest in medical imaging, such as age, pathology or acquisition parameters. It compares favorably with existing methods such as saliency maps. 

The approach may be used for classification, but also provides reconstructions or data generation for each class, which paves the way to a better understanding of class differences. These reconstructions also help to improve the classification accuracy of an independent classifier.

Future  direction  of  this  work  includes comparing the proposed approach with generative adversarial networks that also can achieve disentanglement in a (semi-) supervised setting. 
Moreover, learning the significant differences between the classes (at the population level) during training is another perspective that would help to determine if the differences observed for a particular sample under classification are related to opposite sex reconstruction or if they stem from other reasons such as registration inaccuracy.
This may further help the analysis of the results.

\bibliographystyle{apalike}
\bibliography{mybibfile}

\begin{thebibliography}{}

\bibitem[Adebayo et~al., 2018]{sanityCheck}
Adebayo, J., Gilmer, J., Muelly, M., Goodfellow, I., Hardt, M., and Kim, B.
  (2018).
\newblock Sanity checks for saliency maps.
\newblock In {\em Proceedings of the 32nd International Conference on Neural
  Information Processing Systems}, page 9525–9536.

\bibitem[Agier et~al., 2020]{AGIE2020}
Agier, R., Valette, S., Kéchichian, R., Fanton, L., and Prost, R. (2020).
\newblock Hubless keypoint-based {3D} deformable groupwise registration.
\newblock {\em Medical Image Analysis}, 59.

\bibitem[Arun et~al., 2020]{arun2020assessing}
Arun, N.~T., Gaw, N., Singh, P., Chang, K., Hoebel, K.~V., Patel, J., Gidwani,
  M., and Kalpathy-Cramer, J. (2020).
\newblock Assessing the validity of saliency maps for abnormality localization
  in medical imaging.
\newblock In {\em Medical Imaging with Deep Learning}.

\bibitem[{Bengio} et~al., 2013]{bengio}
{Bengio}, Y., {Courville}, A., and {Vincent}, P. (2013).
\newblock Representation learning: A review and new perspectives.
\newblock {\em IEEE Transactions on Pattern Analysis and Machine Intelligence},
  35(8):1798--1828.

\bibitem[Brůžek et~al., 2017]{BRUZ2017}
Brůžek, J., Santos, F., Dutailly, B., Murail, P., and Cunha, E. (2017).
\newblock Validation and reliability of the sex estimation of the human os
  coxae using freely available {DSP2} software for bioarchaeology and forensic
  anthropology.
\newblock {\em American Journal of Physical Anthropology}, 164(2):440--449.

\bibitem[Chen et~al., 2018]{unsupervised}
Chen, R. T.~Q., Li, X., Grosse, R.~B., and Duvenaud, D.~K. (2018).
\newblock Isolating sources of disentanglement in variational autoencoders.
\newblock In {\em Advances in Neural Information Processing Systems},
  volume~31.

\bibitem[{d’Oliveira Coelho} and Curate, 2019]{CADOES2019}
{d’Oliveira Coelho}, J. and Curate, F. (2019).
\newblock Cadoes: An interactive machine-learning approach for sex estimation
  with the pelvis.
\newblock {\em Forensic Science International}, 302.

\bibitem[Eitel and Ritter, 2019]{eitel2019testing}
Eitel, F. and Ritter, K. (2019).
\newblock Testing the robustness of attribution methods for convolutional
  neural networks in {MRI}-based {A}lzheimer's disease classification.
\newblock In {\em Interpretability of Machine Intelligence in Medical Image
  Computing and Multimodal Learning for Clinical Decision Support}. Springer
  International Publishing.

\bibitem[Erhan et~al., 2009]{vis}
Erhan, D., Bengio, Y., Courville, A., and Vincent, P. (2009).
\newblock Visualizing higher-layer features of a deep network.
\newblock Technical Report 1341, University of Montreal.

\bibitem[Fong et~al., ]{9010039}
Fong, R., Patrick, M., and Vedaldi, A.
\newblock Understanding deep networks via extremal perturbations and smooth
  masks.
\newblock In {\em ICCV}, pages 2950--2958.

\bibitem[Higgins et~al., 2017]{Higgins2017betaVAELB}
Higgins, I., Matthey, L., Pal, A., Burgess, C.~P., Glorot, X., Botvinick, M.,
  Mohamed, S., and Lerchner, A. (2017).
\newblock beta-{VAE}: Learning basic visual concepts with a constrained
  variational framework.
\newblock In {\em ICLR}.

\bibitem[Kingma et~al., 2014]{KingmaD}
Kingma, D., Rezende, D., Mohamed, S., and Welling, M. (2014).
\newblock Semi-supervised learning with deep generative models.
\newblock In {\em Advances in Neural Information Processing Systems}.

\bibitem[Kingma and Welling, 2014]{Kingma2014}
Kingma, D.~P. and Welling, M. (2014).
\newblock {Auto-Encoding Variational {B}ayes}.
\newblock In {\em 2nd International Conference on Learning Representations,
  {ICLR}, Canada, Conference Track Proceedings}.

\bibitem[Komar and Buikstra, 2007]{forensincsAnthropology}
Komar, D. and Buikstra, J. (2007).
\newblock {\em Forensic Anthropology: Contemporary Theory And Practice}.
\newblock Oxford University Press.

\bibitem[Liu et~al., 2020]{mic2}
Liu, R., Subakan, C., Balwani, A.~H., Whitesell, J., Harris, J., Koyejo, S.,
  and Dyer, E.~L. (2020).
\newblock A generative modeling approach for interpreting population-level
  variability in brain structure.
\newblock In {\em Medical Image Computing and Computer Assisted Intervention --
  MICCAI 2020}, pages 257--266. Springer International Publishing.

\bibitem[Murail et~al., 2005]{MURA2005}
Murail, P., Bruzek, J., Houët, F., and Cunha, E. (2005).
\newblock {DSP}: A tool for probabilistic sex diagnosis using worldwide
  variability in hip-bone measurements.
\newblock {\em Bulletins et mémoires de la Société d’Anthropologie de
  Paris}, 17(3-4):167--176.

\bibitem[N et~al., 2017]{Siddhart}
N, S., Paige, B., van~de Meent, J.-W., Desmaison, A., Goodman, N., Kohli, P.,
  Wood, F., and Torr, P. (2017).
\newblock Learning disentangled representations with semi-supervised deep
  generative models.
\newblock In {\em Advances in Neural Information Processing Systems},
  volume~30.

\bibitem[Nguyen et~al., 2019]{Nguyen2019}
Nguyen, A., Yosinski, J., and Clune, J. (2019).
\newblock {\em Understanding Neural Networks via Feature Visualization: A
  Survey}, pages 55--76.
\newblock Springer International Publishing.

\bibitem[Nikita and Nikitas, 2020]{Nikita2020}
Nikita, E. and Nikitas, P. (2020).
\newblock Sex estimation: a comparison of techniques based on binary logistic,
  probit and cumulative probit regression, linear and quadratic discriminant
  analysis, neural networks, and naïve {B}ayes classification using ordinal
  variables.
\newblock {\em International Journal of Legal Medicine}, 134(3):1213--1225.

\bibitem[Ranjan et~al., 2018]{COMA:ECCV18}
Ranjan, A., Bolkart, T., Sanyal, S., and Black, M.~J. (2018).
\newblock Generating {3D} faces using convolutional mesh autoencoders.
\newblock In {\em European Conference on Computer Vision (ECCV)}, pages
  725--741.

\bibitem[Ribeiro et~al., 2016]{LIME}
Ribeiro, M., Singh, S., and Guestrin, C. (2016).
\newblock {``}why should {I} trust you?{''}: Explaining the predictions of any
  classifier.
\newblock In {\em Proceedings of the 2016 Conference of the North {A}merican
  Chapter of the Association for Computational Linguistics: Demonstrations},
  pages 97--101, San Diego, California. Association for Computational
  Linguistics.

\bibitem[Ruiz et~al., 2019]{ruiz2019learning}
Ruiz, A., Martinez, O., Binefa, X., and Verbeek, J. (2019).
\newblock Learning disentangled representations with reference-based
  variational autoencoders.

\bibitem[Rybkin et~al., 2021]{simple}
Rybkin, O., Daniilidis, K., and Levine, S. (2021).
\newblock Simple and effective {VAE} training with calibrated decoders.

\bibitem[Selvaraju et~al., 2019]{Selvaraju_2019}
Selvaraju, R.~R., Cogswell, M., Das, A., Vedantam, R., Parikh, D., and Batra,
  D. (2019).
\newblock Grad-cam: Visual explanations from deep networks via gradient-based
  localization.
\newblock {\em International Journal of Computer Vision}, 128(2):336–359.

\bibitem[Simonyan et~al., 2014]{simonyan2014deep}
Simonyan, K., Vedaldi, A., and Zisserman, A. (2014).
\newblock Deep inside convolutional networks: Visualising image classification
  models and saliency maps.
\newblock In {\em Workshop at International Conference on Learning
  Representations}.

\bibitem[Smilkov et~al., 2017]{smoothgrad}
Smilkov, D., Thorat, N., Kim, B., Vi{\'{e}}gas, F.~B., and Wattenberg, M.
  (2017).
\newblock Smoothgrad: removing noise by adding noise.
\newblock In {\em Workshop on Visualization for Deep Learning, ICML}.

\bibitem[Yan et~al., 2016]{YanYSL16}
Yan, X., Yang, J., Sohn, K., and Lee, H. (2016).
\newblock Attribute2image: Conditional image generation from visual attributes.
\newblock In {\em ECCV (4)}, Lecture Notes in Computer Science, pages 776--791.
  Springer.

\bibitem[Young et~al., 2019]{three}
Young, K., Booth, G., Simpson, B., Dutton, R., and Shrapnel, S. (2019).
\newblock Deep neural network or dermatologist?
\newblock {\em Lecture Notes in Computer Science}.

\bibitem[Zhao et~al., 2019]{mic}
Zhao, Q., Adeli, E., Honnorat, N., Leng, T., and Pohl, K.~M. (2019).
\newblock Variational autoencoder for regression: Application to brain aging
  analysis.
\newblock In {\em Medical Image Computing and Computer Assisted Intervention --
  MICCAI 2019}, pages 823--831. Springer International Publishing.

\bibitem[Zhou et~al., 2016]{CAM2}
Zhou, B., Khosla, A., Lapedriza, A., Oliva, A., and Torralba, A. (2016).
\newblock Learning deep features for discriminative localization.
\newblock In {\em IEEE Conference on Computer Vision and Pattern Recognition
  (CVPR)}.

\end{thebibliography}
\end{document}